\documentclass[conference]{IEEEtran}
\IEEEoverridecommandlockouts
\usepackage[a-1b]{pdfx}
\usepackage{cite}
\usepackage{amsmath,amssymb,amsfonts}
\usepackage{algorithmic}
\usepackage{graphicx}
\usepackage{textcomp}
\usepackage{xcolor}
\usepackage{hyperref}
\usepackage{comment}
\usepackage[caption=false,font=footnotesize]{subfig}
\usepackage{dblfloatfix}
\def\BibTeX{{\rm B\kern-.05em{\sc i\kern-.025em b}\kern-.08em
    T\kern-.1667em\lower.7ex\hbox{E}\kern-.125emX}}
\begin{document}

\title{TimeColor: Flexible Reference Colorization via Temporal Concatenation\\
\thanks{Dr. Hang Su is the corresponding author.}
}

    \author{\IEEEauthorblockN{Bryan Constantine Sadihin}
\IEEEauthorblockA{\textit{Computer Science and Technology} \\
\textit{Tsinghua University}\\
Beijing, China \\
bryan.constantine7@gmail.com}
\and
\IEEEauthorblockN{Yihao Meng}
\IEEEauthorblockA{\textit{Computer Science and Engineering} \\
\textit{HKUST}\\
Hong Kong, China \\
ymengas@connect.ust.hk}
\and
\IEEEauthorblockN{Michael Hua Wang}
\IEEEauthorblockA{\textit{Computer Science and Technology} \\
\textit{Tsinghua University}\\
Beijing, China \\
wanghua24@mails.tsinghua.edu.cn}
\and[\hfill\mbox{}\par\mbox{}\hfill]
\IEEEauthorblockN{Matteo Jiahao Chen}
\IEEEauthorblockA{\textit{Computer Science and Technology} \\
\textit{Tsinghua University}\\
Beijing, China \\
matteonech23@gmail.com}
\and
\IEEEauthorblockN{Hang Su}
\IEEEauthorblockA{\textit{Computer Science and Technology} \\
\textit{Tsinghua University}\\
Beijing, China \\
suhangss@mail.tsinghua.edu.cn}
}

\maketitle

\begin{abstract}
Most colorization models condition only on a single reference, typically the first frame of the scene. However, this approach ignores other sources of conditional data, such as character sheets, background images, or arbitrary colorized frames. We propose TimeColor, a sketch-based video colorization model that supports heterogeneous, variable-count references with the use of explicit per-reference region assignment. TimeColor encodes references as additional latent frames which are concatenated temporally, permitting them to be processed concurrently in each diffusion step while keeping the model's parameter count fixed. TimeColor also uses spatiotemporal correspondence-masked attention to enforce subject--reference binding in addition to modality-disjoint RoPE indexing. These mechanisms mitigate shortcutting and cross-identity palette leakage. Experiments on Sakuga-42M under both single- and multi-reference protocols show that TimeColor improves color fidelity, identity consistency, and temporal stability over prior baselines. Our project page is available at \url{https://bconstantine.github.io/TimeColor/}.
\end{abstract}

\section{Introduction}
\label{sec:intro}

Animation is a cornerstone of contemporary visual media. However, high-quality production remains labor-intensive as modifications must be manually propagated across frames. While manually drawing sketches rewards precise structural control, the colorization process is largely an exercise in constraint enforcement that preserves character identity and palette continuity rather than inventing new content. This is typically achieved through reliance on rich, reusable references, such as character design sheets, background paintings, or colorized frames from earlier shots.

Recent advances treat sketch colorization as a conditional generation problem solved with video diffusion models \cite{Meng2025AniDoc, Yang2025LayerAnimate, Xing2024ToonCrafter, li2025tooncomposer, Huang2024LVCD, Zhang2025AnimeColor, Chen2025LongAnimation, sadihin2025sketchcolour}. Despite progress, existing methods remain limited for production use. Existing colorization methods rely on single-reference conditioning. By tying generation to a single reference sample, additional references cannot be used even when they are available. Furthermore, most prior work requires references drawn from the target shot, typically a colorized version of the first keyframe per cut. This dependency limits cross-scene reuse and increases labor requirements.

As a result, current systems struggle in the scenarios where reference diversity is most valuable, namely where character identity must be maintained across changes in pose and viewpoint while being unable to derive structure from the base sketch. Notably, these cases are more challenging for multi-reference models due to ambiguity regarding which reference should govern which output region, which makes naive conditioning prone to shortcutting or identity leakage.

\begin{figure}[tbp]
\centering
\includegraphics[width=\linewidth]{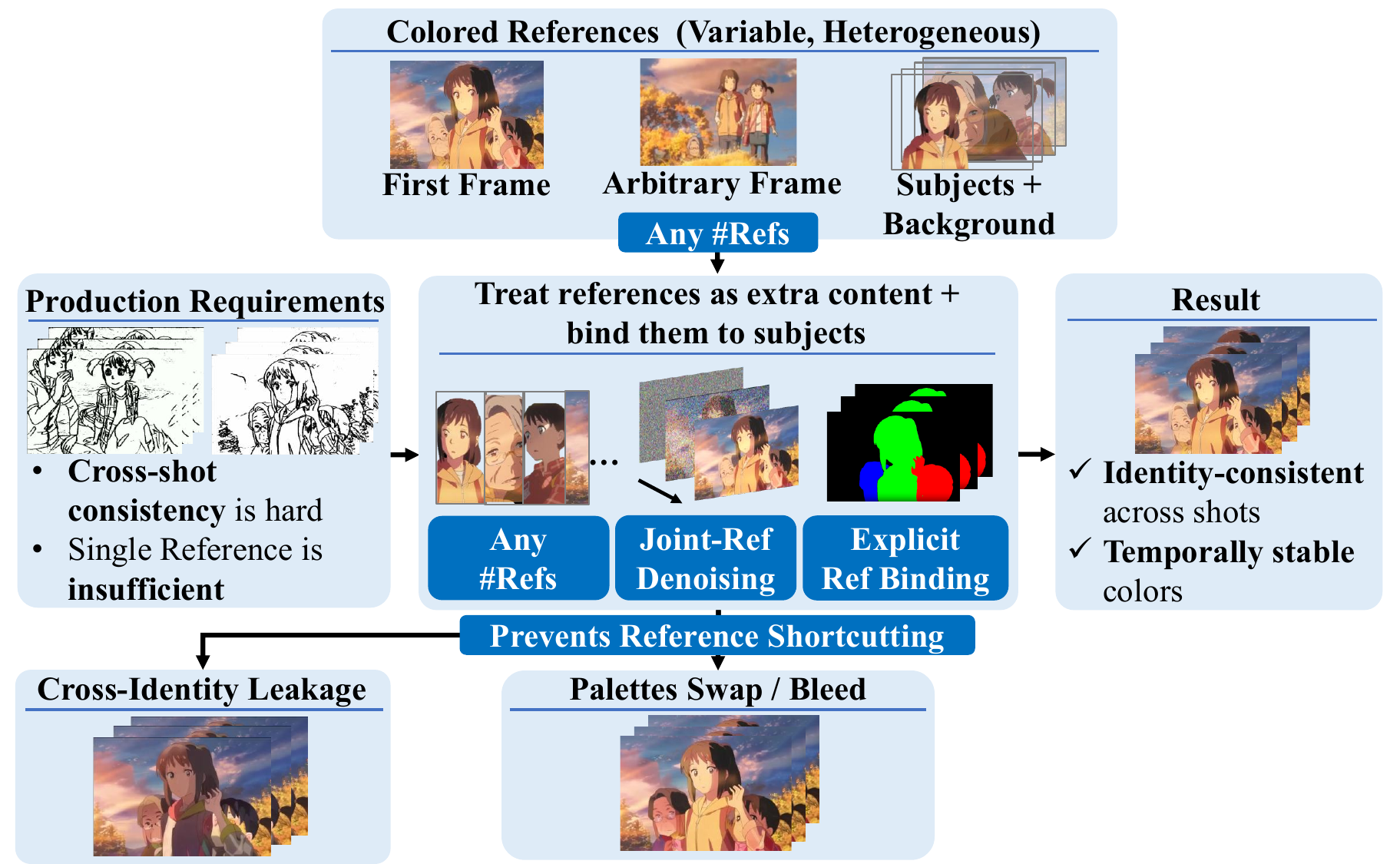}
\caption{\textbf{TimeColor} enables sketch video colorization with a fixed parameter budget, conditioning on heterogeneous, variable-count references. It generates identity-consistent, temporally stable colorized animations from sketch videos, aiming to reduce manual 2D colorization effort.
}
\label{fig:opening_figure}
\end{figure}

We propose TimeColor, a diffusion transformer-based framework for sketch video colorization supporting variable-count, heterogeneous multi-reference conditioning with explicit region-level control. Fig.~\ref{fig:opening_figure} illustrates our overall framework. We design TimeColor around three common production input types: a colorized first frame, a frame from a different timestamp or shot (referred to as arbitrary-frame) and multiple subject/background references (e.g., character sheets). 

Our central idea is to encode all references as additional latent frames injected via temporal concatenation, thus permitting an arbitrary number of references to be processed concurrently in each diffusion step without increasing parameter budget. Importantly, we both apply modality-disjoint RoPE indexing to prevent positional interference across modalities (target, sketch, and reference tokens) and enforce spatiotemporal correspondence-masked attention to bind each subject region to its designated reference set, thus permitting controllable subject--reference assignment. These complementary mechanisms target reference shortcutting and cross-identity leakage, which are the common failure modes in reference-guided generation. This resolves the ambiguity regarding which reference should influence which region.

To train our line art video colorization model, we require a large-scale dataset of cartoons with reliable instance tracking and correspondence. However, such datasets are rare or nonexistent, while manual annotation is prohibitively costly. To overcome this issue, we developed an automated curation pipeline capable of detecting, tracking, and extracting subjects using InternVL3 \cite{Zhu2025InternVL3} and GroundedSAM2 \cite{Liu2024GroundingDINO, Ravi2024SAM2}, allowing us to produce subject/background references and corresponding per-frame dense pixel-level correspondence masks at scale. To increase reference-target appearance gaps, we select same-character references from different scenes when possible using DINO-based retrieval \cite{Maxime2024DinoV2}.

We evaluate TimeColor on the Sakuga-42M test set \cite{sakuga42m2024} under diverse reference regimes, including starting-frame, arbitrary-frame, and multi-reference settings. TimeColor improves color fidelity, identity preservation, and temporal stability across tested settings over prior methods, with the largest gains in the multi-reference regime, where reference ambiguity and leakage are most pronounced.


Our contributions are as follows:
(1) We propose TimeColor, a DiT-based framework for sketch video colorization with heterogeneous, variable-count references through temporal sequence conditioning.
(2) We propose modality-disjoint RoPE with correspondence-masked attention to enforce subject--reference binding and mitigate reference shortcutting for controllable multi-reference colorization.
(3) We introduce an automated pipeline that constructs large-scale multi-reference sketch video colorization data, including subject/background references and pixel-level correspondence masks.

\section{Related Work}
\label{sec:related_work}

\subsection{Controllable Video Generation}
Controllable video diffusion introduces additional conditioning signals to provide finer control over the generated video result. However, common integration methods such as ControlNet-style adapter branches \cite{zhang2023controlnet} or channel-wise feature injection \cite{Saharia2023SuperResolution} typically assume a fixed number and layout of control inputs, making them unsuitable for variable-count reference sets. In contrast,  diffusion transformer (DiT) \cite{peebles2023dit} models video as spatiotemporal token sequence, allowing conditioning to be appended as additional latent frames \cite{Jiang2025VACE} supporting elastic conditioning length. We study variable-count reference conditioning, where per-reference region mapping is enforced concurrently during denoising.

\subsection{Sketch Colorization}
Diffusion-based approaches have improved reference adherence for sketch colorization. Recent sketch image-level colorization systems have extended reference guidance to multiple references \cite{Zhang2025MagicColor, Zhuang2025Cobra}. At the same time, video diffusion models are adopted for sketch video colorization to improve temporal consistency. However, existing work relies on a single colored reference, most require it to originate from the target shot. These methods differ in how they inject sketch/reference signals. LVCD uses ControlNet to condition on a previously colored frame \cite{Huang2024LVCD}. ToonCrafter conditions on colored endpoints for colorization/interpolation \cite{Xing2024ToonCrafter}. AniDoc leverages point-map to explore correspondence with character sheets \cite{Meng2025AniDoc}. Newer methods use DiT to pursue finer control. AnimeColor conditions on a single image using ControlNet and feature injection \cite{Zhang2025AnimeColor}. LongAnimation introduces long-range generation with global–local memory from a starting-frame reference \cite{Chen2025LongAnimation}. ToonComposer unifies in-betweening and colorization via sparse-sketch injection \cite{li2025tooncomposer}. LayerAnimate decomposes single-reference into motion-aware layers \cite{Yang2025LayerAnimate}. 

Concurrent with our work, InstanceAnimator \cite{anonymous2025instanceanimator} explores instance-level conditioning for sketched video colorization. In contrast, we use hard spatiotemporal attention masking to constrain reference colorization to the intended video regions.

\section{Method}
\label{sec:method}

\begin{figure*}[tbp]
\centering
\includegraphics[width=\linewidth]{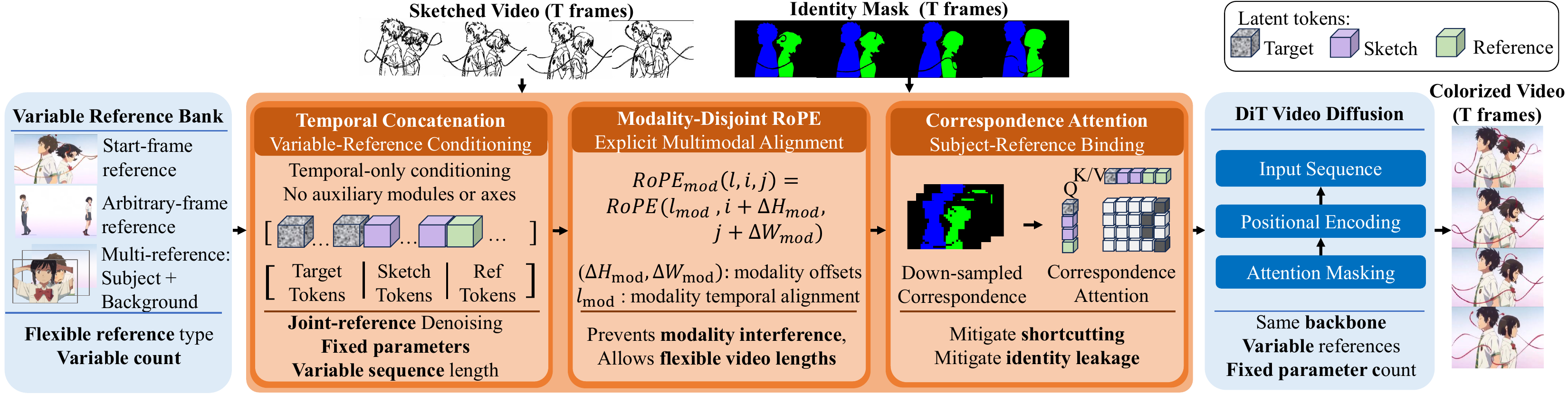}
\caption{\textbf{Overview of TimeColor}. Given a sketched video and a variable-length reference bank (starting-frame, arbitrary-frame, and multi-reference cues), TimeColor conditions a DiT video diffusion model via temporal token concatenation, modality-disjoint RoPE, and correspondence-masked attention to bind subjects to references while mitigating shortcutting/identity leakage, enabling flexible reference types and lengths with a fixed backbone and parameter count.
\vspace{-3ex}
}
\label{fig:method_overview}
\end{figure*}

We study diffusion-based sketch video colorization with spatially grounded multi-reference conditioning, where a variable number of heterogeneous references require region-level control to mitigate cross-identity color leakage. Fig.~\ref{fig:method_overview} illustrates TimeColor, a DiT-based video diffusion model that encodes conditioning signals as additional latent “frames” and injects them via temporal concatenation, enabling variable-count multi-reference conditioning with a fixed parameter budget. A hard correspondence-aware attention mask further enforces explicit subject--reference binding.

\subsection{Problem Formulation}
\label{subsec:problem_setup}

Given a sketch video $S=\{S_t\}_{t=1}^{T}$ and a reference set $I_{\mathrm{ref}}=\{I_r\}_{r=1}^{R}$, each reference $I_r$ can be a colored starting-frame, an arbitrary (possibly cross-shot) frame, or a subject/background sheet. We additionally assume mutually exclusive correspondence masks
$\mathcal{M}_{\mathrm{ref}}=\{\mathcal{M}_{t,r}\}_{t=1}^{T}{}_{r=1}^{R}$, where $\mathcal{M}_{t,r}(x,y)\in\{0,1\}$ assigns each pixel $(x,y)$ in frame $t$ to exactly one reference index $r$.\footnote{Background can be treated as an additional reference index when applicable.}
Our goal is to generate a colorized video $Y=\{Y_t\}_{t=1}^{T}$ that satisfies palette fidelity to the assigned references, identity-consistent colors across subjects, and temporal coherence between frames.

Let $z_n$ denote the noisy latent at diffusion step $n$. Let $\mathcal{C}$ denote the conditioning bundle, consisting of sketches $S$, references $I_{\mathrm{ref}}$, correspondence masks $\mathcal{M}_{\mathrm{ref}}$, and optional text $c$. We train with the standard noise-prediction objective:
\begin{equation}
\mathcal{L}_{\mathrm{noise}}
=\mathbb{E}_{z_0,n,\epsilon}\!\left[\left\|\epsilon-\epsilon_{\theta}(z_n, n, \mathcal{C})\right\|_2^2\right],
\end{equation}
where $\epsilon\sim\mathcal{N}(0,I)$ and $n$ are sampled uniformly. Loss supervision is applied only on target latents as sketches/references act purely as conditioning signals.

\subsection{Temporal Concatenation for Variable-Count Conditioning}
\label{subsec:temporal_concatenation}

\begin{figure}[!t]
  \centering
  \includegraphics[width=\linewidth]{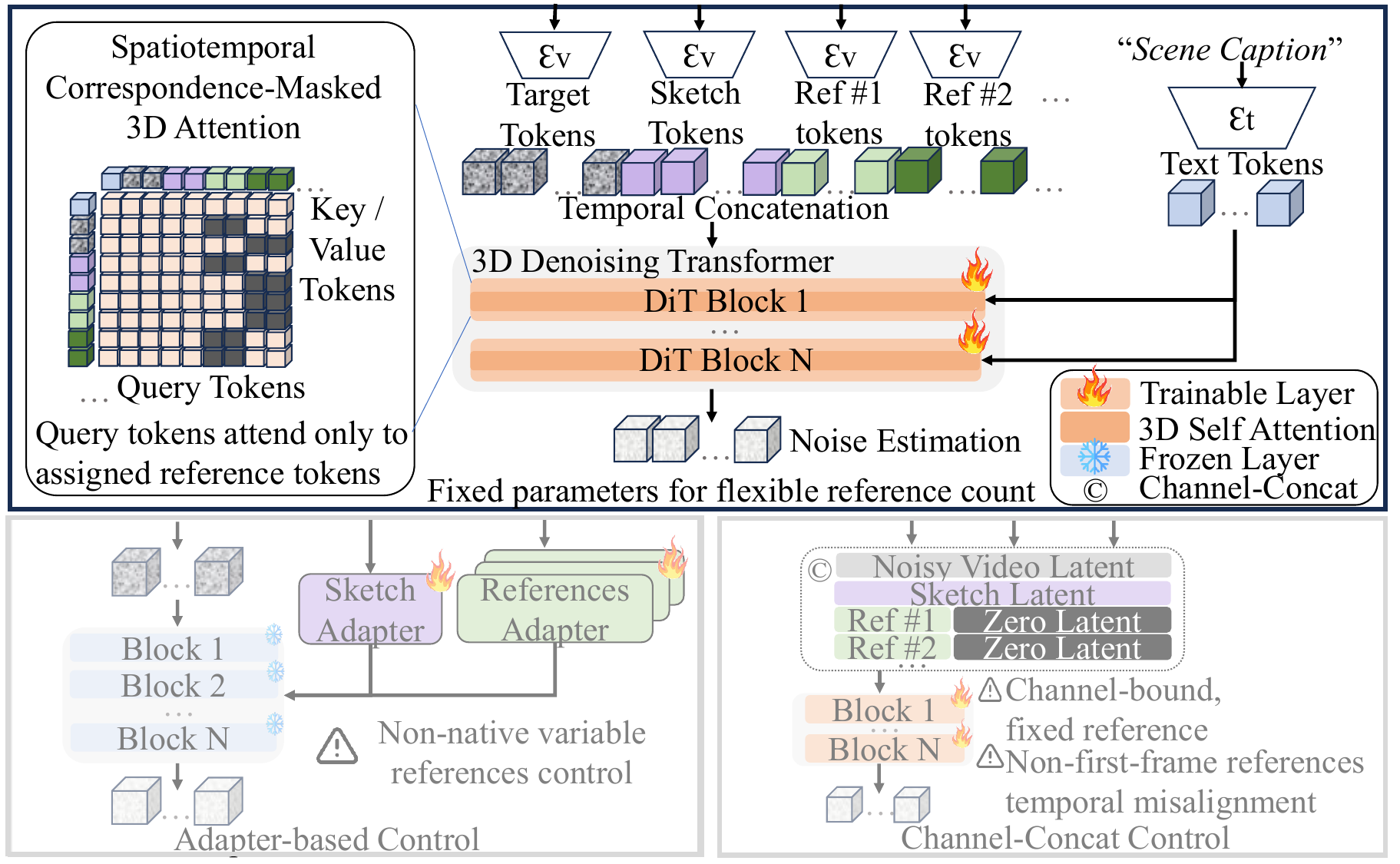}
  \caption{\textbf{TimeColor model diagram}. Target video, sketch, and variable image reference tokens are temporally concatenated. Correspondence-masked attention restricts attention to the assigned reference, enforcing strict reference correspondence. Unlike adapter/channel-stacking controls, TimeColor enables concurrent variable-count reference conditioning with fixed parameters and robustness to non-starting-frame misalignment.}
  \label{fig:model_diagram}
  \vspace{-2.5ex}
\end{figure}

Existing video colorization methods inject references via channel stacking \cite{sadihin2025sketchcolour, li2025tooncomposer, Chen2025LongAnimation} or control branches/adapters \cite{Meng2025AniDoc, Xing2024ToonCrafter, Zhang2025AnimeColor, Chen2025LongAnimation, Yang2025LayerAnimate}, which either assume a fixed number of reference channels or require multi-pass inference as references vary. As illustrated in Fig.~\ref{fig:model_diagram}, TimeColor instead offloads the conditioning fully to the temporal dimension of DiT \cite{peebles2023dit}: embedding the sketch video and each reference image into full-resolution token grids (at the same latent resolution as the target video) and concatenating them along the temporal axis. This yields concurrent reference conditioning for an arbitrary number of references. Increasing $R$ changes only sequence length (compute), not parameters. As shown in Sec.~\ref{sec:experiments}, common workarounds degrade colorization quality. Collaging references into a single image \cite{Meng2025AniDoc} reduces per-reference token coverage. Two-step “colorize starting-frame then propagate video” pipelines or multi-pass inference per-reference conditioning accumulate errors while increasing inference cost.

\paragraph{Modality-disjoint RoPE}
Naively concatenating heterogeneous modalities can cause positional interference. To maintain flexible video length at inference time, we assign disjoint RoPE \cite{Su2024RoFormer} index ranges across modalities. For a token from modality $m\in\{0,1,2\}$ with temporal index $l$ and spatial indices $(i,j)$, we apply
\begin{equation}
\mathrm{RoPE}_{m}(l,i,j)
= \mathrm{RoPE}\!\big(l,\; i+mH,\; j+mW\big),
\end{equation}
where $m{=}0/1/2$ correspond to (noised target, sketch, reference) tokens and $H,W$ are offsets ensuring non-overlap. Noised target and sketch tokens share the same frame index $l$ to preserve frame alignment. Reference images are assigned distinct negative indices $l=-r$ ($r\ge 1$), keeping them non-overlapping and temporally separated.

\subsection{Spatiotemporal Correspondence-Masked Attention}
\label{subsec:masked_attention}

A core failure mode in multi-reference conditioning is shortcutting. The model attends to whichever reference looks most similar, which induces cross-identity palette leakage. As shown in Fig.~\ref{fig:model_diagram}, DiT applies 3D self-attention over spatiotemporal patch tokens, so reference tokens can mix unless explicitly constrained. We show that VAE downsampling preserves coarse layout cues (see supplementary material) and that tokenization operates locally on patches. Since these operations preserve patch-level structure, we enforce a hard correspondence mask that assigns each spatiotemporal location to exactly one reference.

\paragraph{Latent-level correspondence IDs}
Given pixel-level one-hot masks $\{\mathcal{M}_{t,r}\}_{r=1}^{R}$, we downsample them to the DiT spatiotemporal patch grid using the same spatiotemporal strides as tokenization and assign each target/sketch patch $i$ a reference identity $\rho(i)$ via pooled majority voting. For reference tokens, we set $\rho(i)=r\in\{1,\dots,R\}$.

\paragraph{Attention gating}
Let $\pi(i)\in\{\textsc{text},\textsc{target},\textsc{sketch},\textsc{ref}\}$ denote the modality of token $i$, and let $\rho(i)\in\{1,\dots,R\}$ be its assigned reference identity. We construct a binary attention mask $M_{ij}\in\{0,1\}$ indicating whether query token $i$ is allowed to attend to key/value token $j$:
\begin{equation}
\begin{aligned}
M_{ij} &= \mathbb{I}\!\left[\pi(i)=\textsc{text}\ \lor \pi(j)\neq\textsc{ref}\ \lor\ \rho(j)=\rho(i)\right],\\
\alpha_{ij}&= \mathrm{softmax}_j\!\left(\frac{q_i^\top k_j}{\sqrt d} + (1-M_{ij})\cdot(-\infty)\right),\\
o_i &= \sum_j \alpha_{ij} v_j,
\end{aligned}
\end{equation}
where $\mathbb{I}[\cdot]$ is the indicator function. Thus, all tokens may attend to non-reference tokens, while attention to reference tokens is permitted only within the same identity ($\rho(j)=\rho(i)$), mitigating cross-reference mixing. This reduces shortcutting and cross-identity palette leakage while retaining single-pass inference. Alternative designs, such as concatenating masks as an additional conditioning stream, are discussed in the supplementary materials. Soft conditioning can easily be ignored by the model and fail to enforce correspondence.

\subsection{Multi-Reference Tracking Dataset Generation}
\label{subsec:multiref_curation}
Animation datasets with per-subject multi-reference tracking are scarce but crucial for reference-conditioned video training. Building supervision at scale is challenging due to 2D instance tracking under varying pose/occlusion, the need for non-starting-frame references to avoid model copying from near-duplicate references (reference-shortcut), and reference sampling that avoids missing/irrelevant references.

We scale the creation of a heterogeneous reference-conditioned animation colorization dataset with an automated pipeline. For each scene, we obtain instance tracks and per-frame pixel-level correspondence by first enumerating main subjects in the scene with InternVL3~\cite{Zhu2025InternVL3}, then grounding text queries on sampled keyframes using Grounding DINO~\cite{Liu2024GroundingDINO}, and finally propagating instance masks over time with SAM2~\cite{Ravi2024SAM2}. The propagated masks define which pixels correspond to each reference instance for each ground-truth frame.
To handle occlusions and late-appearing subjects, we run this procedure iteratively over multiple keyframes and merge only newly discovered instances at each pass.

For each scene of length \(L\), we incorporate reference-sampling gap $g$ to increase reference-target appearance gaps. Specifically, we supervise on the last \(f\) frames (RGB target), generate corresponding binarized sketches following \cite{Meng2025AniDoc, Chen2025LongAnimation} and sample references from a source window spanning \([1,\,L-g-f]\).
We extract three reference settings: starting-frame (the first RGB frame of the supervision window), arbitrary-frame (a single RGB frame sampled from the source window), and multi-reference (a set of per-instance RGB references plus one background reference). In multi-reference, we keep only instances that remain visible across the supervision window and exceed a minimum area. For each, we choose the source-window frame with the largest mask (as a proxy for minimum occlusion), crop the corresponding RGB region and increase reference diversity via augmentations (center-crop/horizontal flip/resize) plus, when available, DINO-based cross-scene retrieval within the same video~\cite{Maxime2024DinoV2}. The background reference is sampled from the source window with all selected objects masked out to mitigate inter-reference leakage. Our dataset pipeline is illustrated in the supplementary material.

\section{Experiments}
\label{sec:experiments}
\subsection{Implementation Details}
\label{sec:implementation-details}
We build on CogVideoX-5B \cite{yang2025cogvideox} at \(480\times720\) resolution. We run our automated reference-generation pipeline (Sec.~\ref{subsec:multiref_curation}) on the Sakuga-42M training set~\cite{sakuga42m2024} with multi-subject tag, which is scene-split and captioned. We set reference-sampling gap $g=17$. This yields \(\sim\)120K starting-frame/arbitrary-frame and \(\sim\)96K multi-reference samples. We train with AdamW \cite{Ilya2019AdamW} (lr \(=1\times10^{-5}\)) using a three-stage curriculum that progressively increases conditioning difficulty: starting-frame to arbitrary-frame to multi-reference for 20K update steps each on 6 A40 GPUs. Further implementation details are provided in the supplementary material.

\subsection{Main Results}
\label{subsection:comparison}

\begin{figure*}[tbp]
\centering

\subfloat[Starting-frame reference]{
  \includegraphics[width=0.24\linewidth]{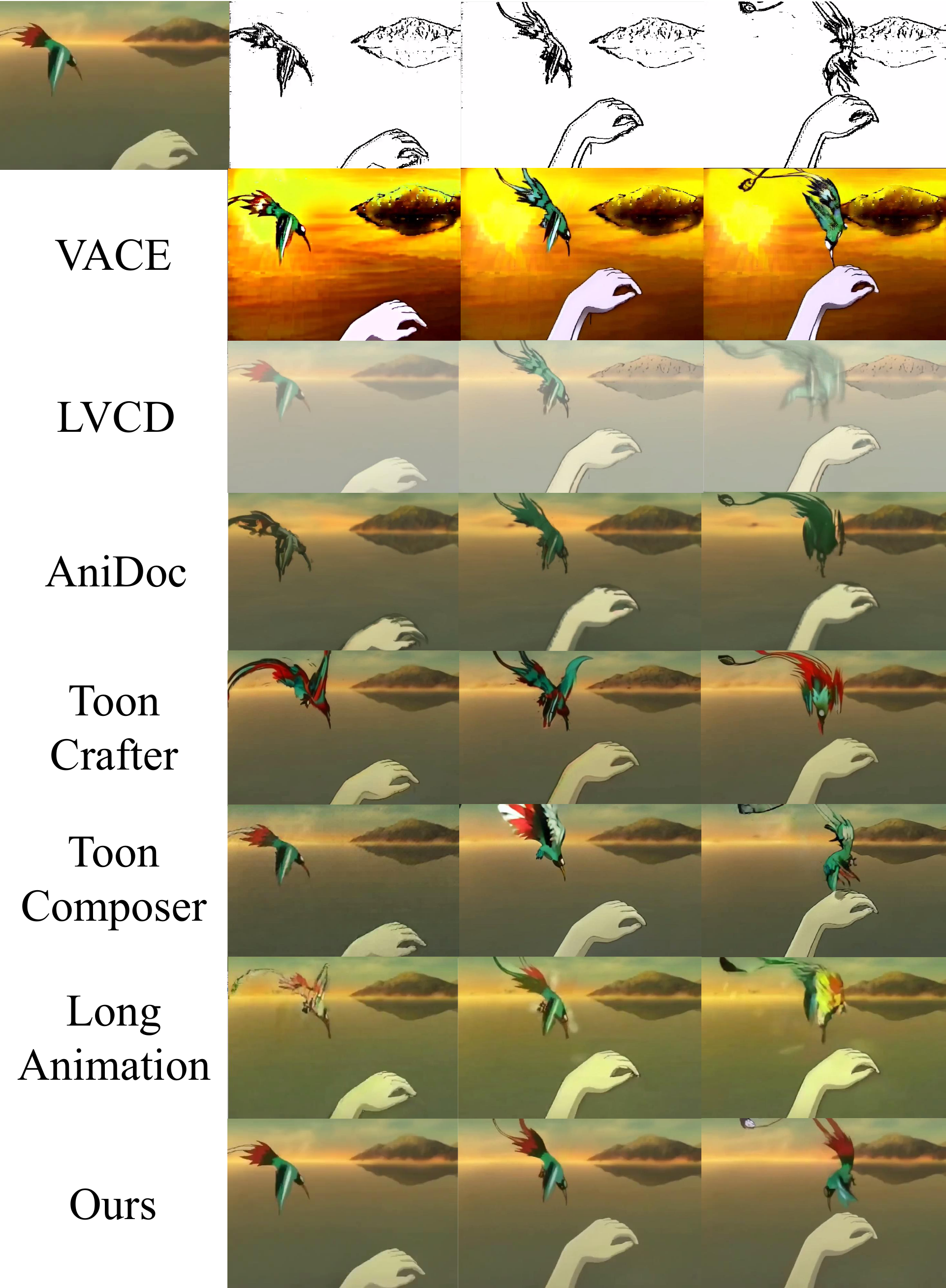}
  \label{fig:stage0_qual}
}
\subfloat[Arbitrary-frame reference]{
  \includegraphics[width=0.24\linewidth]{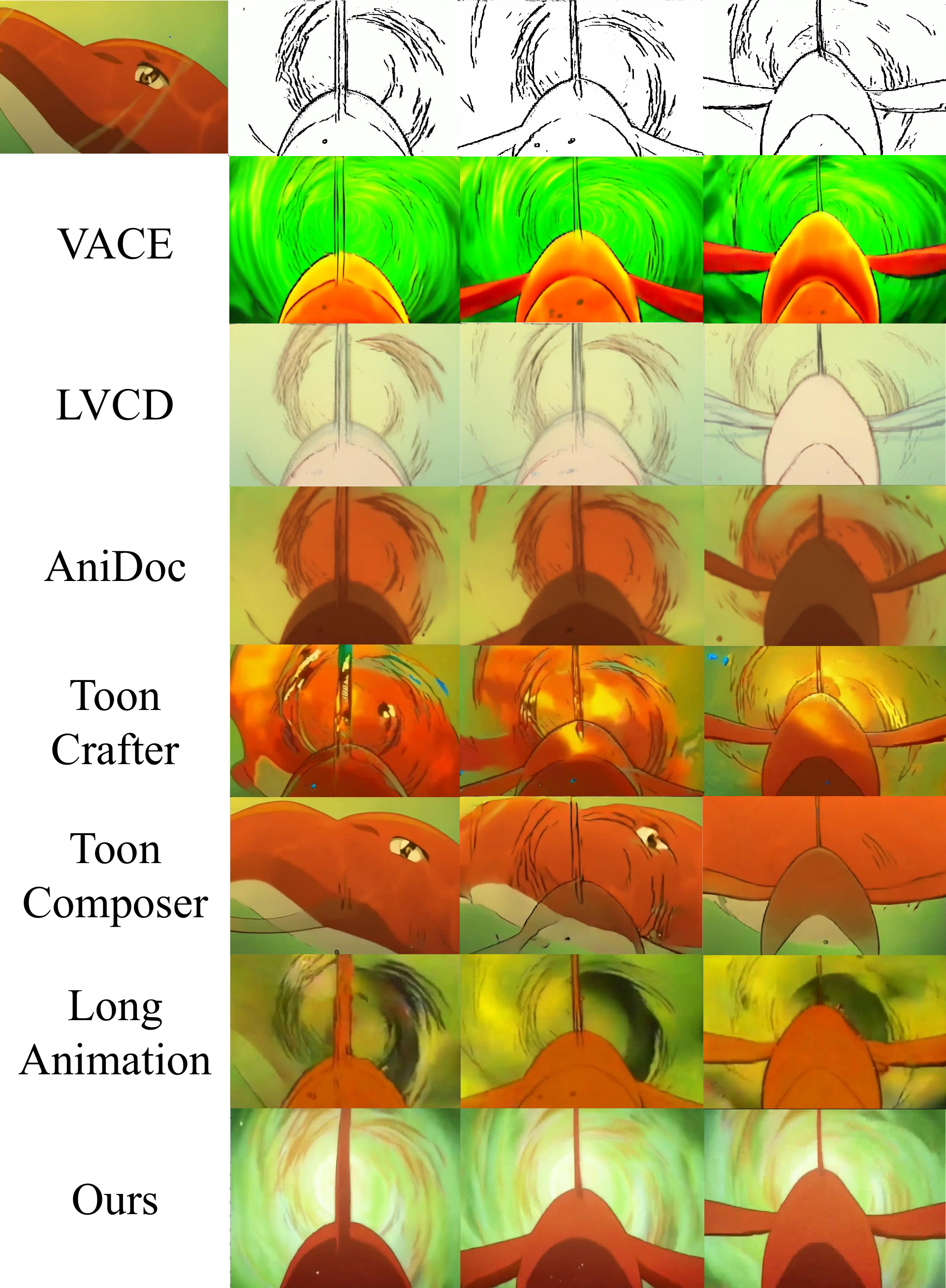}
  \label{fig:stage1_qual}
}
\subfloat[Multi-reference 1]{
  \includegraphics[width=0.24\linewidth]{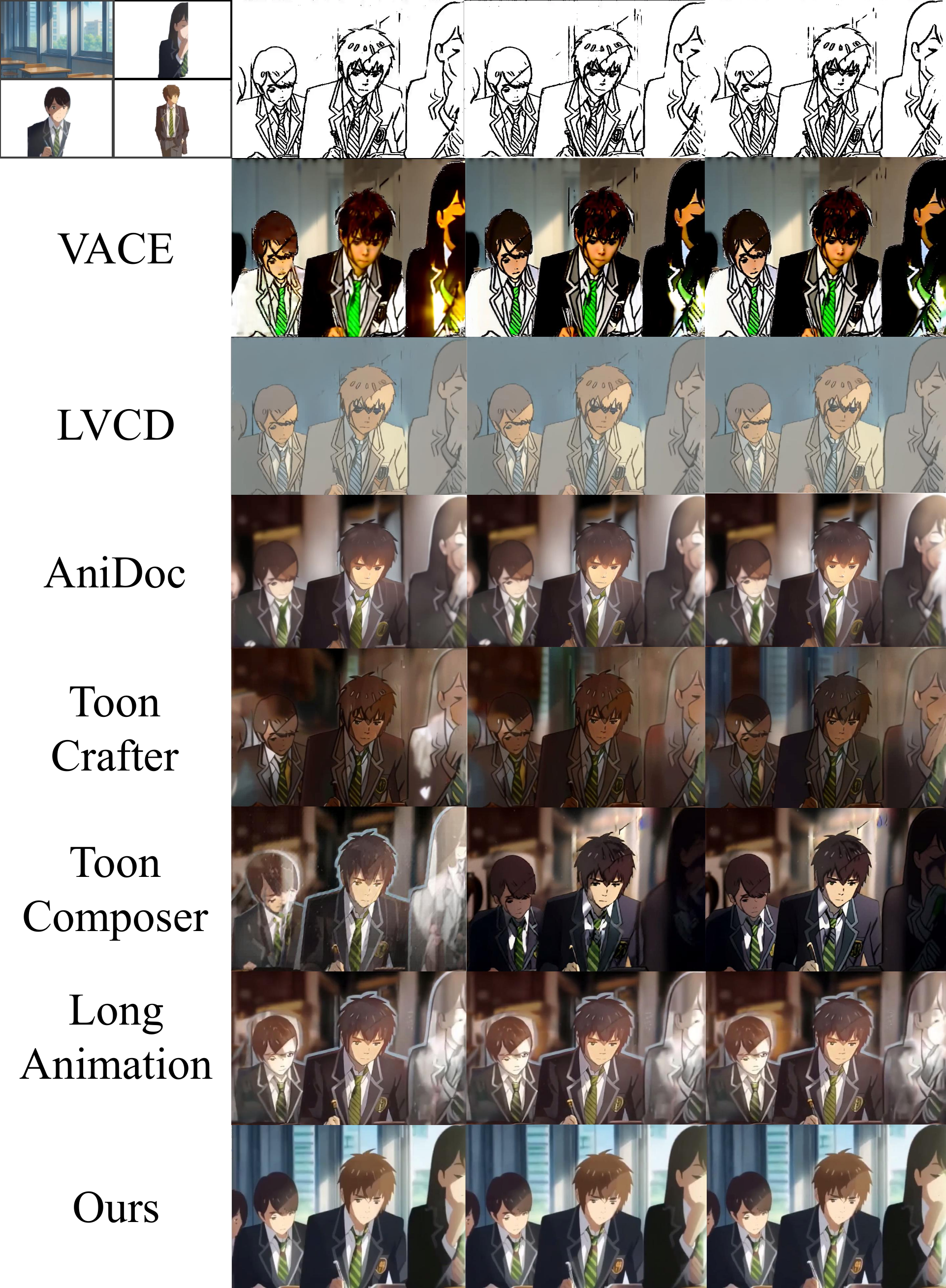}
  \label{fig:stage2_qual_2}
}
\subfloat[Multi-reference 2]{
  \includegraphics[width=0.24\linewidth]{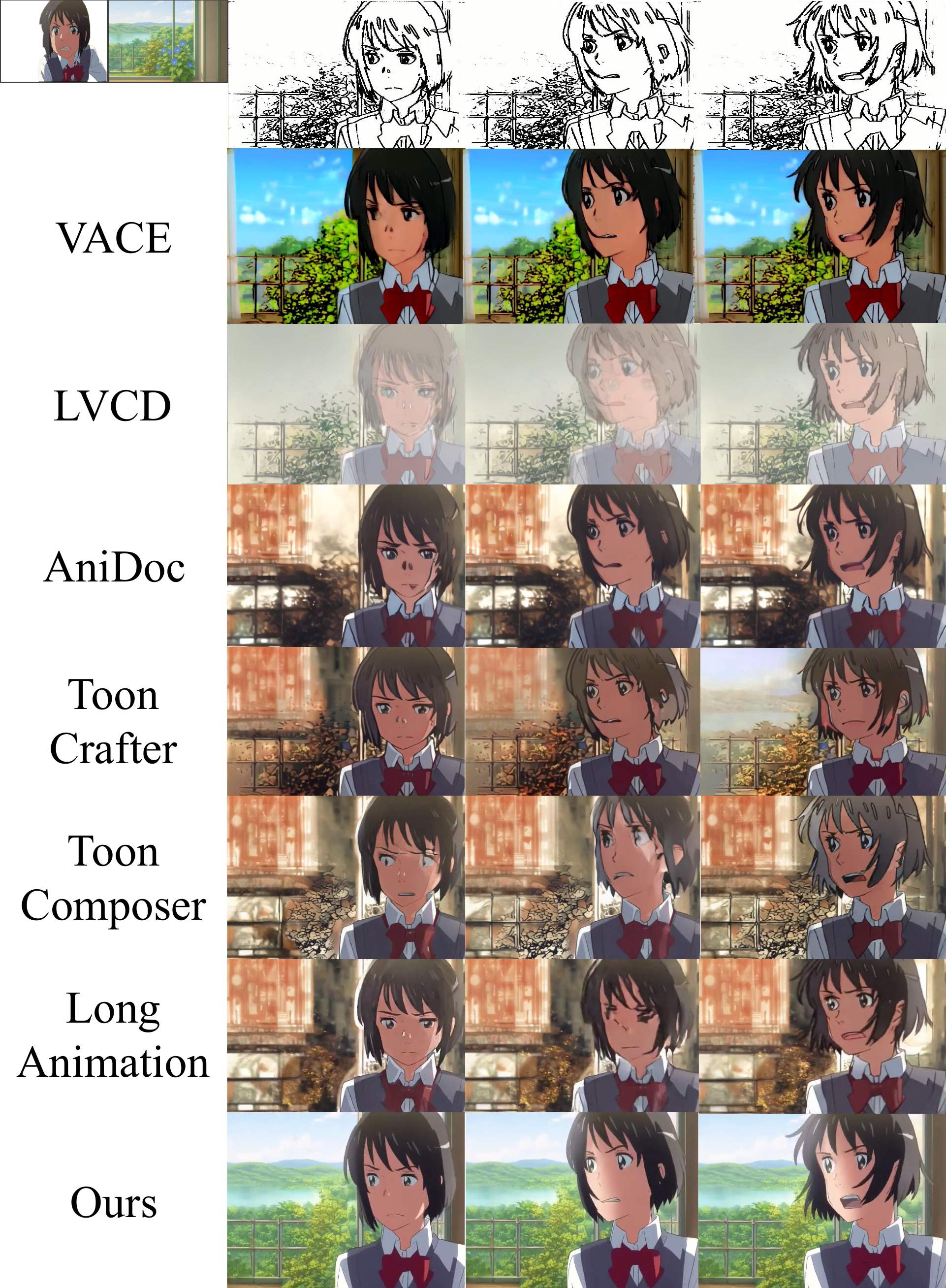}
  \label{fig:stage2_qual_1}
}

\caption{Qualitative comparison among baselines under single-reference (starting-frame and arbitrary-frame) and multi-reference settings across seven methods: VACE \cite{Jiang2025VACE}, LVCD \cite{Huang2024LVCD}, AniDoc \cite{Meng2025AniDoc}, ToonCrafter \cite{Xing2024ToonCrafter}, ToonComposer \cite{li2025tooncomposer}, LongAnimation \cite{Chen2025LongAnimation}, and TimeColor.}
\label{fig:qualitative_comparison}
\vspace{-2ex}
\end{figure*}

We conduct evaluations on the Sakuga-42M test split with multi-subject tag. Using our multi-reference curation pipeline (Sec.~\ref{subsec:multiref_curation}), we generate per-clip references and retain only samples with verified reference--video correspondence, yielding \(\sim\)1{,}200 evaluation clips. Masked-out backgrounds are inpainted with Nano Banana to approximate production reference inputs. We then compare against state-of-the-art animation video colorization open-source baselines that utilize channel concatenation or adapter-based conditioning, each run at their native clip lengths: LVCD \cite{Huang2024LVCD}, AniDoc \cite{Meng2025AniDoc}, ToonCrafter \cite{Xing2024ToonCrafter} (with the last colored frame included), ToonComposer \cite{li2025tooncomposer}, and LongAnimation \cite{Chen2025LongAnimation}. Furthermore, we evaluate against VACE \cite{Jiang2025VACE} as a general video-to-video diffusion editor that supports colorization and multi-reference conditioning. For fairness, we evaluate the first 14 frames for all methods, since some baselines natively support only 14-frame clips. Because baselines accept only a single colored reference, for multi-reference evaluation, we use a two-step protocol: Cobra \cite{Zhuang2025Cobra} first colorizes the starting-frame using the multi-reference inputs, and the resulting image is used as the reference for video colorization. For ToonComposer, dense-sketch conditioning is noisy. We therefore follow the official demo’s maximum of four sketches and sample them uniformly across the evaluated clip (see supplementary material). In ToonComposer’s arbitrary-frame setting, we re-index so the colored reference is frame 0 and apply the same temporal offset to sketches.

Additional baseline workarounds are reported in the supplementary material: (i) tiling multi-reference images into a single grid input (as noted in \cite{Meng2025AniDoc}), (ii) a two-step arbitrary-frame baseline where Cobra colorizes the first frame and (iii) multi-pass colorization per reference for VACE with per-reference masking, all of which perform worse.

\paragraph{Quantitative Comparison.} Following prior work \cite{Meng2025AniDoc, Chen2025LongAnimation}, we resize all videos/frames to 256$\times$256 and report FID~\cite{Heusel2017FID} for frame-distribution quality, FVD~\cite{Unterthiner2019FVD} for video-level quality, and PSNR, LPIPS~\cite{Zhang2018LPIPS}, and SSIM~\cite{Wang2004SSIM} for frame-wise similarity. Table~\ref{tab:quant_stage012} shows that TimeColor achieves the best score across all settings. Notably, in the harder arbitrary-frame and multi-reference regimes, TimeColor remains competitive with baselines evaluated under the simpler starting-frame condition, indicating robustness to reference diversity. All results use the same model trained with our three-stage curriculum. To reflect annotation flexibility, we additionally evaluate a starting-frame-only mask setting where masks are drawn on the first frame and propagated from sketches using SAM2 \cite{Ravi2024SAM2} (with mIoU=0.803 against test masks), under which TimeColor largely preserves performance.

\begin{table}[tbp]
\setlength{\tabcolsep}{3pt}
\caption{\textbf{Quantitative comparison with prior baselines across settings.} TimeColor uses the same weights across settings. \textit{Prop. Masks}: starting-frame masks propagated from sketches. \textbf{Bolded}: Best, \underline{underlined}: Second best. }
\label{tab:quant_stage012}
\centering
\begin{tabular}{@{}lrrrrr@{}}
\hline
Method & SSIM $\uparrow$ & PSNR $\uparrow$ & LPIPS $\downarrow$ & FVD $\downarrow$ & FID $\downarrow$ \\
\hline
\multicolumn{6}{c}{Single Reference, Starting-Frame} \\
\hline
VACE \cite{Jiang2025VACE}   & 0.4810 & 12.85 & 0.4018 & 385.05 & 51.60 \\
LVCD \cite{Huang2024LVCD}   & 0.5469 & 11.18 & 0.3996 & 288.19 & 30.52 \\
AniDoc \cite{Meng2025AniDoc} & 0.7536 & 20.79 & 0.2133 & 63.44 & 20.67 \\
ToonCrafter   \cite{Xing2024ToonCrafter}               & 0.7487 & 21.75 & 0.1895 & 66.83 & 14.38 \\
ToonComposer  \cite{li2025tooncomposer}               & 0.7046 & 20.09 & 0.2371 & 67.54 & 12.78\\
LongAnimation   \cite{Chen2025LongAnimation}            & 0.7193 & 20.34 & 0.2461 & 71.51 & 16.51 \\
\textbf{TimeColor (Ours)}                & \textbf{0.8496} & \textbf{24.95} & \textbf{0.1309} & \textbf{30.74} & \textbf{9.44} \\
\hline
\multicolumn{6}{c}{Single Reference, Arbitrary-Frame} \\
\hline
VACE \cite{Jiang2025VACE}   & 0.4600 & 12.24 & 0.4238 & 396.19 & 53.78 \\
LVCD  \cite{Huang2024LVCD}   & 0.5189 & 10.49 & 0.4436 & 344.27 & 39.60 \\
AniDoc \cite{Meng2025AniDoc}  & 0.7189 & 18.97 & 0.2555 & 81.27 & 24.14 \\
ToonCrafter \cite{Xing2024ToonCrafter}  & 0.6957 & 19.47 & 0.2415 & 93.32 & 18.17 \\
ToonComposer \cite{li2025tooncomposer}                 & 0.5657 & 15.31 & 0.3611 & 138.04 & 22.05 \\
LongAnimation \cite{Chen2025LongAnimation}                & 0.6592 & 18.04 & 0.3105 & 102.36 & 21.96 \\
\textbf{TimeColor (Ours)}                & \textbf{0.8071} & \textbf{21.98} & \textbf{0.1822} & \textbf{37.68} & \textbf{11.99} \\
\hline
\multicolumn{6}{c}{Multi-Reference} \\
\hline
VACE \cite{Jiang2025VACE}   & 0.3369 & 9.76 & 0.5342 & 454.40 & 64.01 \\
LVCD  \cite{Huang2024LVCD}   & 0.4846 & 10.58 & 0.5198 & 426.89 & 57.11 \\
AniDoc \cite{Meng2025AniDoc}  & 0.5798 & 13.50 & 0.4042 & 181.01 & 43.78 \\
ToonCrafter \cite{Xing2024ToonCrafter}  & 0.5002 & 13.02 & 0.4173 & 199.23 & 43.65 \\
ToonComposer \cite{li2025tooncomposer}                 & 0.4294 & 12.00 & 0.5135 & 157.35 & 30.71 \\
LongAnimation \cite{Chen2025LongAnimation}                & 0.4731 & 12.68 & 0.4841 & 200.44 & 38.58 \\
\textbf{TimeColor (Ours)} & \textbf{0.7589} & \underline{18.89} & \textbf{0.2361} & \textbf{47.13} & \textbf{15.62} \\
\textbf{TimeColor (Prop. Masks)} & \underline{0.7585} & \textbf{18.95} & \underline{0.2364} & \underline{49.21} & \underline{15.72} \\
\hline
\end{tabular}
\end{table}

\begin{figure}[tbp]
\centering
\vspace{-0.75ex}
\includegraphics[width=\linewidth]{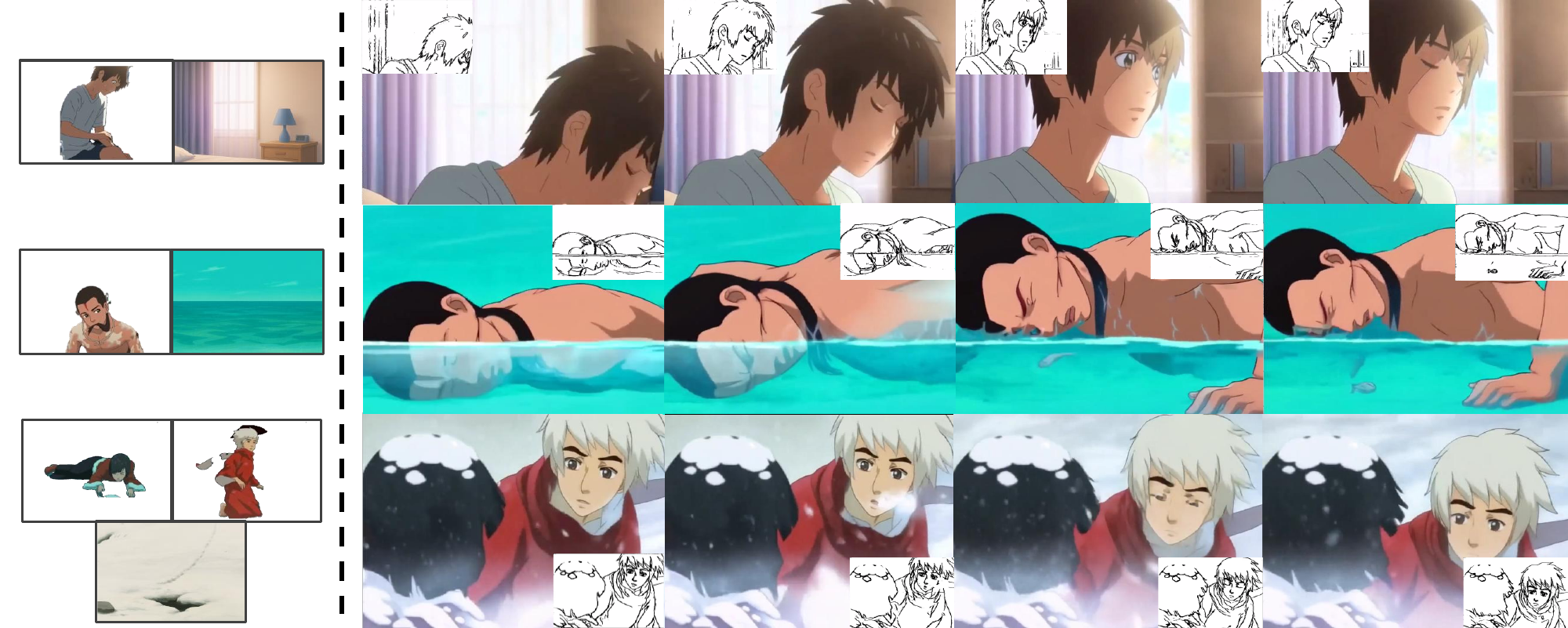}
\caption{\textbf{Robustness to mismatched viewpoints.}
With large pose/viewpoint gaps between references and targets, TimeColor maintains temporal coherence and palette fidelity while avoiding cross-reference leakage.}
\label{fig:different_viewpoint}
\vspace{-0.25ex}
\end{figure}

\paragraph{Qualitative Comparison} As shown in Fig.~\ref{fig:qualitative_comparison}, across starting-frame, arbitrary-frame and multi-reference settings, TimeColor produces colorization that better follows references and exhibits stronger temporal consistency. With a starting-frame reference, our method better preserves palette fidelity and edge adherence, whereas baselines often show desaturation or color bleeding. Under arbitrary-frame reference, baselines frequently mis-map colors or ignore the reference, while our results maintain the intended subject-background palette and structure. In multi-reference scenarios (Fig.~\ref{fig:qualitative_comparison} and Fig.~\ref{fig:different_viewpoint}), TimeColor's spatiotemporal correspondence-masked attention explicitly maps references to target regions while maintaining colorization and subject motion. Additional qualitative results are provided in the supplementary material, including (i) reference reuse and (ii) swapping references between subjects.

\subsection{Ablation Study}
\label{subsection:ablation_study}

We ablate TimeColor by removing modules and evaluating all variants under identical settings (Table~\ref{tab:quant_ablation}). We train a starting-frame reference model for 20K update steps with and without modality-disjoint RoPE. Without modality-disjoint RoPE, sketch and noised target tokens become entangled under temporal concatenation, leading to washed-out colors in later frames and instability beyond the training length. We also compare three fully trained variants: full attention, masking only among reference tokens (preventing each reference token from attending to other references), and our spatiotemporal correspondence-masked attention. Full attention exhibits cross-reference palette interference. While reference-to-reference masking reduces this leakage, it still induces spurious subject-reference associations when colored references differ substantially from the target sketch and multiple subjects share similar cues (such as hairstyles). In contrast, spatiotemporal correspondence masking improves reference mapping robustness, especially under heterogeneous and mismatched references. Additional ablation visualizations are provided in the supplementary material.

\begin{table}[tbp]
\setlength{\tabcolsep}{3pt}
\caption{\textbf{Ablations across Design Components.} \textbf{Bolded}: Best.}
\label{tab:quant_ablation}
\centering
\begin{tabular}{@{}lrrrrr@{}}
\hline
Method & SSIM $\uparrow$ & PSNR $\uparrow$ & LPIPS $\downarrow$ & FVD $\downarrow$ & FID $\downarrow$  \\
\hline
\multicolumn{6}{c}{Single Reference, Starting-Frame} \\
\hline
w/o Custom RoPE              & 0.7042 & 20.63 & 0.2432 & 46.18 & 9.65 \\
\textbf{Custom RoPE}     & \textbf{0.8478} & \textbf{24.81} & \textbf{0.1344} & \textbf{38.50} & \textbf{8.94} \\
\hline
\multicolumn{6}{c}{ Multi-Reference} \\
\hline
Full Attention   & 0.7004 & 16.51 & 0.2878 & 68.95 & 19.49 \\
Mask Inter-Reference Query    & 0.7322 & 18.04 & 0.2543 & 52.20 &16.09 \\
\textbf{Mask Correspondence}     & \textbf{0.7589} & \textbf{18.89} & \textbf{0.2361} & \textbf{47.13} & \textbf{15.62} \\
\hline
\end{tabular}
\vspace{-1.5ex}
\end{table}




\section{Conclusion and Future Work}

We present TimeColor, a diffusion transformer framework for sketch-based video colorization that conditions on heterogeneous references of variable count via temporal concatenation, while keeping parameter count fixed. Modality-disjoint RoPE and spatiotemporal correspondence-masked attention preserve subject--reference binding under multi-reference inputs. On Sakuga-42M, TimeColor outperforms prior work in color fidelity and temporal coherence across single- and multi-reference settings. Future work will relax our correspondence assumptions by reducing reliance on dense masks and instead leveraging sparse correspondence cues.

\section*{Acknowledgment}
This project is supported by  NSF of China Projects (Nos. 92370124, U25B6003, 62550004, 92248303); Beijing Natural Science Foundation L247011 and the Shandong Provincial Natural Science Foundation No. ZR2022ZD01 and the High Performance Computing Center, Tsinghua University.

\bibliographystyle{IEEEtran}
\bibliography{icme2026references_originalwording_cited_shorterconf}

\end{document}


\title{Supplementary Material for TimeColor: Flexible Reference Colorization via Temporal Concatenation
}

\maketitle

\appendix
\subsection{Flexible Usage}
\label{subsection:flexible_usage}
\begin{figure}[tbp]
\centering
\includegraphics[width=\linewidth]{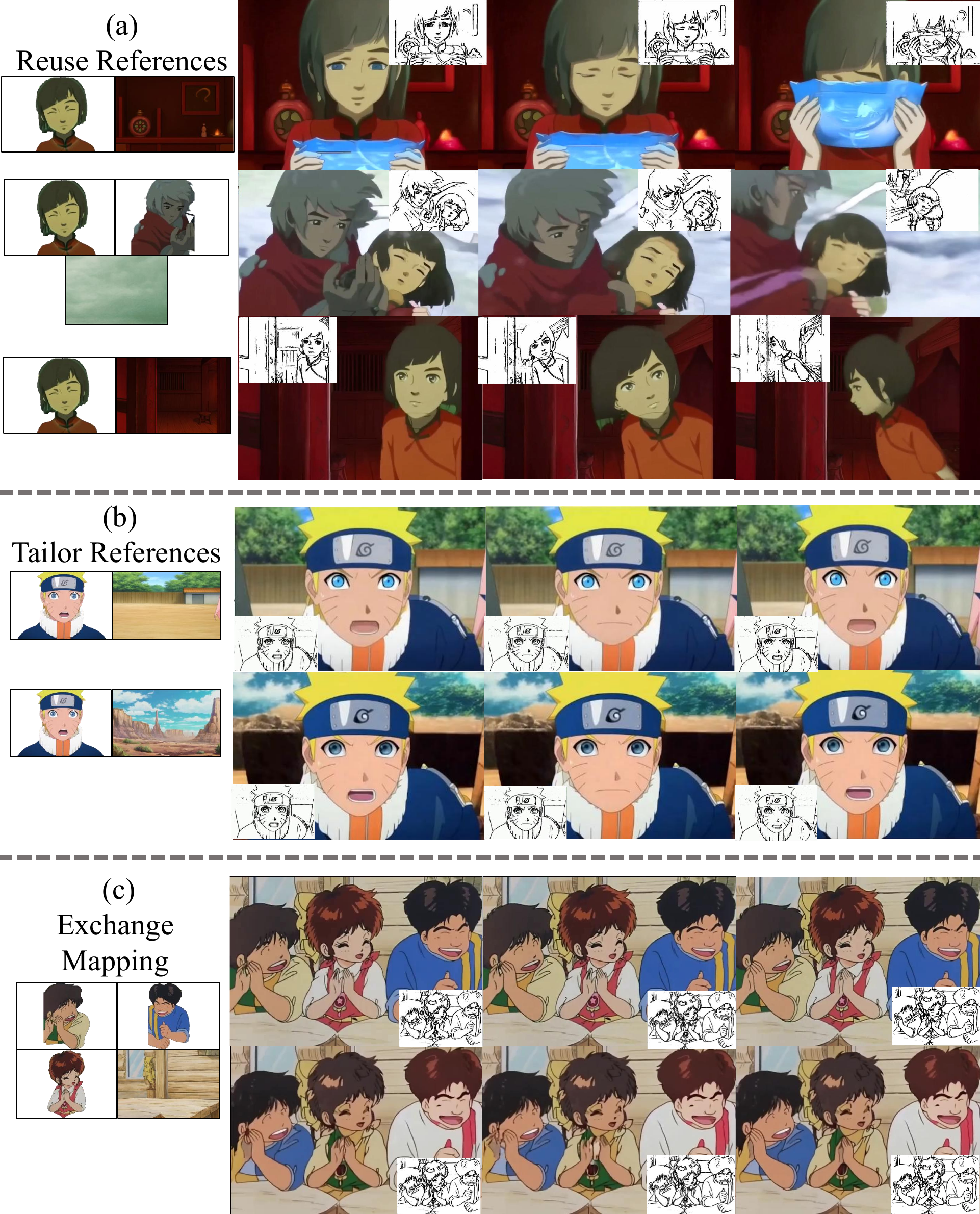}
\caption{\textbf{Flexible usage.} (a) A shared reference can be used across subjects in different scenes. (b) Replacing references steers the colorization appearance. (c) References can be swapped to exchange the subject–reference mapping.}
\label{fig:flexible_usage}
\end{figure}

TimeColor supports reference-based conditioning for diverse production workflows. As shown in Fig.~\ref{fig:flexible_usage}, TimeColor can reuse the same reference for different target sketches, steer the generated output by substituting an alternative reference, and swap the subject--reference assignments. The illustration shows that the model follows explicit subject-to-reference mapping rather than relying on implicit similarity matching.

\subsection{Dataset Pipeline Details}
\label{app:dataset_diagram}

\begin{figure*}[t]
  \centering
          \includegraphics[width=\textwidth]{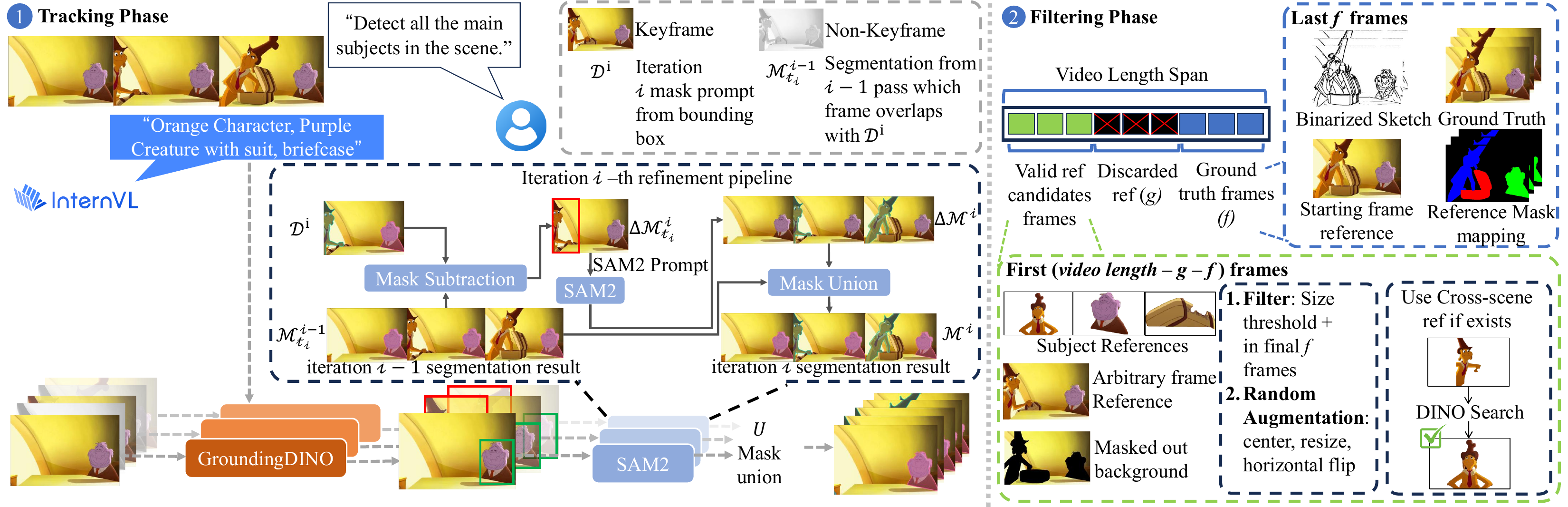}
  \caption{\textbf{Multi-reference dataset pipeline.}
(1) \emph{Tracking:} InternVL3 proposes main subjects, Grounding DINO detects them on keyframes, SAM2 propagates/refines masks. We iterate over keyframes and only propagate newly discovered instances to reduce misses.
(2) \emph{Filtering:} references are sampled from a prefix window to enforce reference-target appearance mismatch. Subjects must satisfy visibility/area constraints in the supervised suffix. Background references are obtained by masking selected subjects. Diversity is increased by random augmentations and cross-scene DINO retrieval when available.}
  \label{fig:dataset_curation}
\end{figure*}

Our dataset creation pipeline details are shown in Fig.~\ref{fig:dataset_curation}. It receives a single scene animation video as input and tracks main subjects throughout the video. This is done by InternVL3 enumerating the video’s main subjects in text, followed by Grounding DINO detecting objects on keyframes from this text list and SAM2 tracking masks across the video. Because a single keyframe can underrepresent or omit subjects, we expand coverage with iterative passes.

\paragraph{Iterative refinement.}
A single-frame source for Grounding DINO inference is prone to underrepresentation of main subjects, including subject occlusion or non-appearance, which can cause Grounding DINO to miss seed detections for propagation. We mitigate this with iterative refinement. Let $V_t$ denote the $t$-th frame, and let keyframes $\{K^i\}_{i=1}^{k_*}$ be sampled every $h{=}5$ frames with frame indices $t_i$.
Let $\mathcal{M}_t^{\,i}$ be the set of propagated masks at frame $t$ after the $i$-th pass, and $\mathcal{M}^{i}$ the union across all frames after pass $i$. Let $\mathcal{O}$ be the InternVL-extracted object list.
At pass $i$, we obtain detections
$D^i = \mathrm{SAM2}(\mathrm{GroundingDINO}(K^i,\mathcal{O}),\,V)$
and keep only elements unseen in the previous pass:
\begin{equation}
\Delta \mathcal{M}_{t_i}^{\,i} \;=\; D^i \setminus \mathcal{M}_{t_i}^{\,i-1}.
\end{equation}
When $\Delta \mathcal{M}_{t_i}^{\,i}\neq\varnothing$, the mask is propagated throughout the video with SAM2
$\Delta \mathcal{M}^{\,i}=\mathrm{SAM2}(\Delta \mathcal{M}_{t_i}^{\,i},\,V_{t_i:\mathrm{end}})$
and update
\begin{equation}
\mathcal{M}^{\,i} \;=\; \mathcal{M}^{\,i-1} \cup \Delta \mathcal{M}^{\,i}.
\end{equation}
After the final pass $k_*$, $\mathcal{M}^{\,k_*}$ covers the tracked masks for all discovered objects.

\paragraph{Reference Filtering and Sketch Generation}
In multi-reference inference, the model must tolerate reference-target mismatches in viewpoint, proportion, and scale. Therefore,  reference-sketch pairs are generated accordingly. Let \(L\) be the length of a single-scene cartoon video sample, \(f\) be the supervision window (the last \(f\) frames) and \(g\) the minimum gap between the earliest reference and the window start. Ground-truth RGBs are the last \(f\) frames. We retain an instance only if (i) it remains visible throughout the last \(f\) frames, (ii) it also appears in \([1,\,L-g-f]\), hereafter referred to as source window, and (iii) its pixel area exceeds a threshold. For object references, we pick the frame within the source window with maximal area. For background, we sample a frame in the source window and remove all selected objects to mitigate leakage. Object and background references are mutually exclusive. Following AniDoc and LongAnimation, we binarize training and test sketches to avoid color leakage. Inputs are the last \(f\) sketches and the ground-truth target is the colored video.

To further increase reference-target appearance diversity, we apply probabilistic augmentations to extracted reference such as recentering, resizing references to the frame size, and performing horizontal flips. When available, DINO embeddings are used to mine cross-scene references of the same instance within the same video.

\subsection{Ablation Study Figures}
\label{app:Ablation_study_figures}

\begin{figure}[tbp]
\centering
\includegraphics[width=\linewidth]{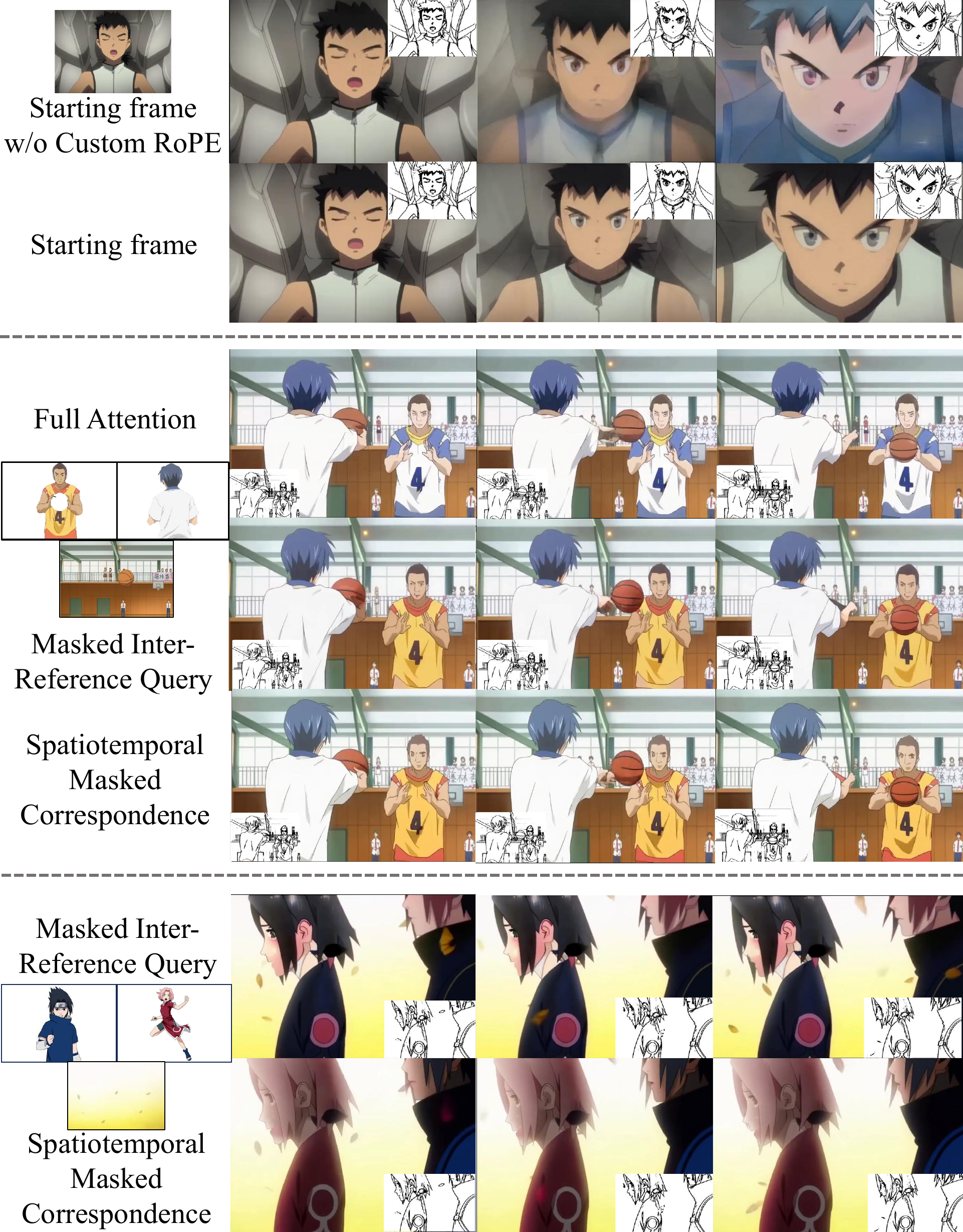}
\caption{\textbf{Ablations of key components}. We ablate model performance with and without RoPE modification, and fully trained model with three different attention variants: full attention, masking only among reference tokens, and spatiotemporal masked correspondence.}
\label{fig:qualitative_ablation}
\end{figure}

We provide qualitative ablation visualizations in Fig.~\ref{fig:qualitative_ablation}, corresponding to the quantitative results of the ablation in the main section. We compare modality-disjoint RoPE and attention-masking strategies under identical training and inference settings. When modality-disjoint RoPE is not applied, later frames exhibit washed-out colors due to entanglement between token modalities. In the multi-reference setting, cross-reference interference occurs between subjects. While reference-to-reference masking reduces this leakage, palette swapping between references can still occur when subjects differ in viewpoint and share similar cues, such as subject hairstyles.

\subsection{Additional Implementation Details}
\label{sec:implementation-details-appendix}

\paragraph{Model and data-generation specifics.} We build on CogVideoX-5B (DiT) at \(480\times720\) resolution. Balancing training temporal context and GPU utilization, we set supervision window \(f{=}17\) and a minimum frame gap \(g{=}17\) for our automated data-generation pipeline. From Sakuga-42M, we select videos tagged \emph{multi-subject} and treat the final 17 frames as ground truth. This pipeline yields around 120K samples for single-reference and full-frame-with-gap settings, and 96K valid multi-reference samples.

\paragraph{Training and inference.} Experiments are run on \(6\times\) NVIDIA A40 using FSDP with batch size of 3, where we apply gradient accumulation over 2 steps (effective global batch size 6). We train with AdamW (lr \(=1\times10^{-5}\)) using a three-stage curriculum that progressively increases conditioning difficulty: starting-frame to arbitrary-frame to multi-reference. Each stage is trained for 20K update steps on 6 A40 GPUs. (\(\sim\)7 days with FSDP). Inference is conducted on a single NVIDIA A40.

\subsection{TimeColor Evaluation with Imperfect Masks}
\label{app:ImperfectMask}
We evaluate TimeColor under imperfect instance masks to better reflect practical deployment, where annotations may be coarse and per-frame masks are often unavailable. Since masks are used only to build our hard spatiotemporal correspondence constraint, this experiment tests sensitivity to mask noise. Starting from the original test masks, we construct: (i) morphed masks by randomly dilating/eroding each subject mask by 5--8 pixels, and (ii) propagated masks by annotating only the first frame and using SAM2 to propagate masks across the clip conditioned on the sketch frames. We report the mean IoU to the original masks as a corruption indicator. As shown in Table~\ref{tab:ImperfectMaskQuantitative}, TimeColor remains stable under both perturbations, with only minor variations across metrics.

\begin{table}
\setlength{\tabcolsep}{2pt}
\caption{\textbf{Quantitative comparison of TimeColor under imperfect instance masks.}
We perturb masks via random dilation/erosion (\emph{Morphed}, 5--8 px) or starting-frame-only annotation with SAM2 propagation (\emph{Propagated}). IoU to original masks is reported in parentheses.}
\centering
\label{tab:ImperfectMaskQuantitative}
\begin{tabular}{@{}lrrrrr@{}}
\hline
Mask processing & SSIM $\uparrow$ & PSNR $\uparrow$ & LPIPS $\downarrow$ & FVD $\downarrow$ & FID $\downarrow$ \\
\hline
\multicolumn{6}{c}{Multi-Reference} \\
\hline
Original Mask  & 0.7589 & 18.89 & 0.2361 & 47.13 & 15.62  \\
Morphed (mean IoU = 0.840)  & 0.7570 & 18.79 & 0.2383 & 47.93 & 15.78 \\
Propagated (mean IoU = 0.803) & 0.7585 & 18.95 & 0.2364 & 49.21 & 15.72 \\
\hline
\end{tabular}
\end{table}

\subsection{Reference Shortcutting under Mask-as-Condition Temporal Concatenation}
\label{subsection:reference_shortcutting}
\begin{figure}[tbp]
\centering
\includegraphics[width=\linewidth]{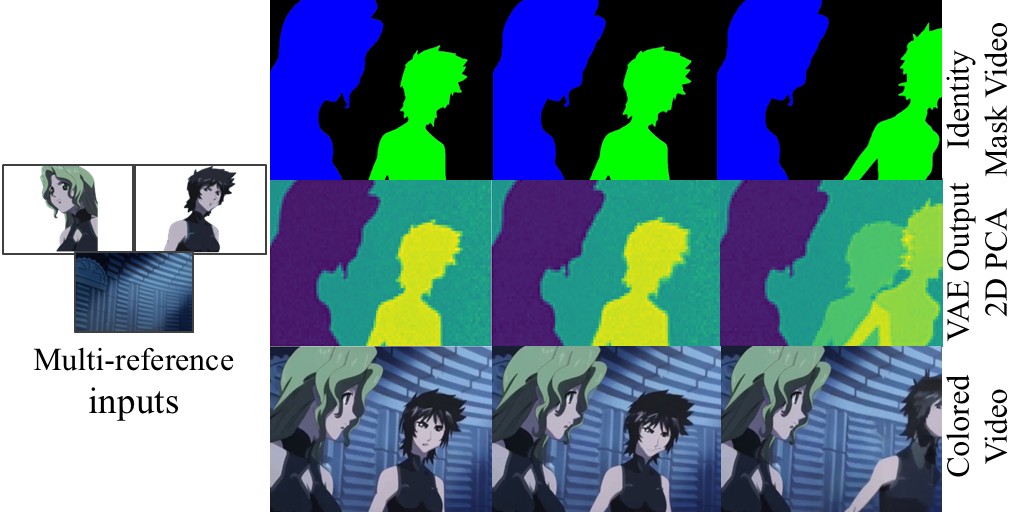}
\caption{\textbf{VAE-encoded colored correspondence masks.} We encode per-pixel reference assignments as an RGB-coded mask video (each reference ID mapped to a distinct color) and pass it through the same VAE encoder as other visual inputs. PCA visualization of the resulting mask latents (projecting along the channel dimension) shows that different reference colors remain separable and spatially coherent after spatiotemporal compression, suggesting the correspondence signal is preserved in latent space.}
\label{fig:colored_mask_compact}
\end{figure}

As a first attempt to impose spatiotemporal correspondence without explicit attention gating, we explored mask-as-conditioning under the same temporal-concatenation architecture used for multi-reference training. Concretely, we concatenated a correspondence signal as an additional conditional stream: given mutually exclusive per-pixel assignments, we encode the reference index using a colored mask (each reference is mapped to a distinct RGB code in image space) and pass it through the same VAE encoder as other visual inputs. To verify that this signal is not trivially destroyed by compression, we analyze the resulting mask latents by applying PCA along the channel dimension, and observe that different reference colors remain clearly separable in the VAE latent space and remain coherent (see Fig.~\ref{fig:colored_mask_compact}). This suggests the VAE preserves sufficient information for the model to, in principle, recover reference identity and spatial assignment.

\begin{figure}[tbp]
\centering
\includegraphics[width=\linewidth]{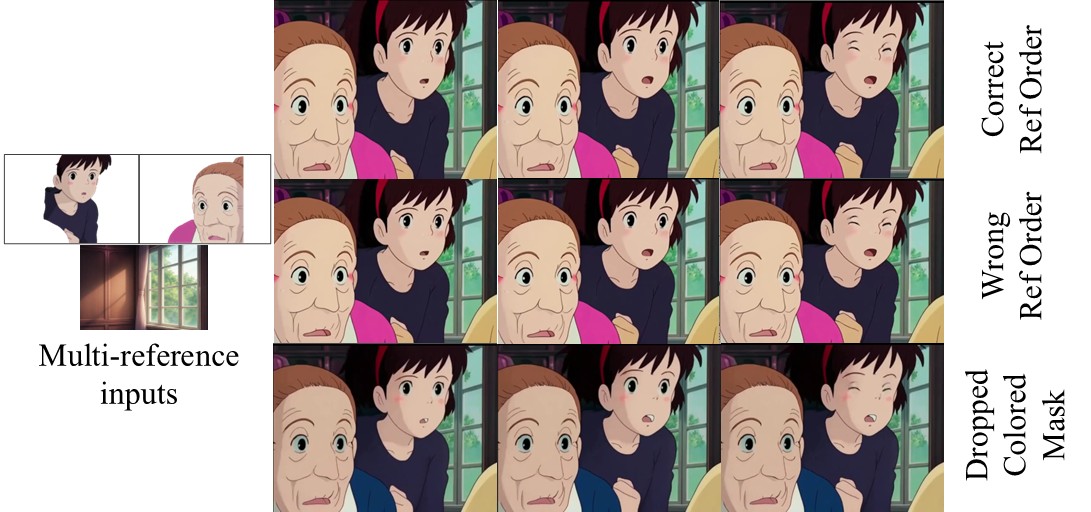}
\caption{\textbf{Order insensitivity under mask-as-conditioning.} We train a full-attention variant that temporally concatenates a VAE-encoded colored correspondence mask during the multi-reference stage. Despite the mask being separable in latent space, the model remains insensitive to reference ordering and fails to maintain consistent subject--reference association, indicating it can ignore the soft correspondence cue and rely on a similarity-based shortcut.}
\label{fig:order_insensitivity}
\end{figure}

Motivated by this observation, we train a variant that uses full attention and temporally concatenates a VAE-encoded colored correspondence mask during the multi-reference stage, while keeping the remaining training protocol and compute budget identical to the main model. As shown in Fig.~\ref{fig:order_insensitivity}, although the mask remains distinguishable in latent space, the trained model is insensitive to reference ordering and fails to preserve a stable subject–reference binding. Moreover, ablating the mask at inference time only causes localized color degradation (e.g., partial desaturation or minor palette drift) rather than a global failure, suggesting the mask is not a core dependency of the generation process. Instead, the model exploits a shortcut: it matches each sketch region to the most visually similar reference (by sketch/shape cues), effectively ignoring the explicit correspondence signal. This behavior indicates that under temporal concatenation, a soft correspondence cue is easily treated as optional conditioning, and optimization can converge to an easier solution that does not learn reliable region-to-reference binding.

This “reference shortcut” phenomenon motivates our hard mask gating design: instead of providing correspondence as an additional input that can be ignored, we enforce it at the mechanism level by restricting attention so that each target region can attend only to its assigned reference tokens. Hard gating turns correspondence from a hint into a constraint, mitigating cross-reference leakage and order-invariant shortcut that arise when correspondence is injected purely as a concatenated conditional.

\subsection{Alternative VACE Evaluation via Multi-Pass Inference with Mask}
\label{app:MultipassInferenceVACE}
We experimented colorization with VACE that imitates our per-reference mask mapping. As VACE masked editing only supports binary mask, we emulate the same per-reference mapping constraint with a multi-pass protocol. Specifically, we run one masked edit per reference: for each reference, we convert its correspondence assignment into a binary mask video, apply VACE to edit only the masked region conditioned on the corresponding image reference, and composite the edited region into an evolving canvas video used for subsequent passes.

\begin{figure}[tbp]
\centering
\includegraphics[width=\linewidth]{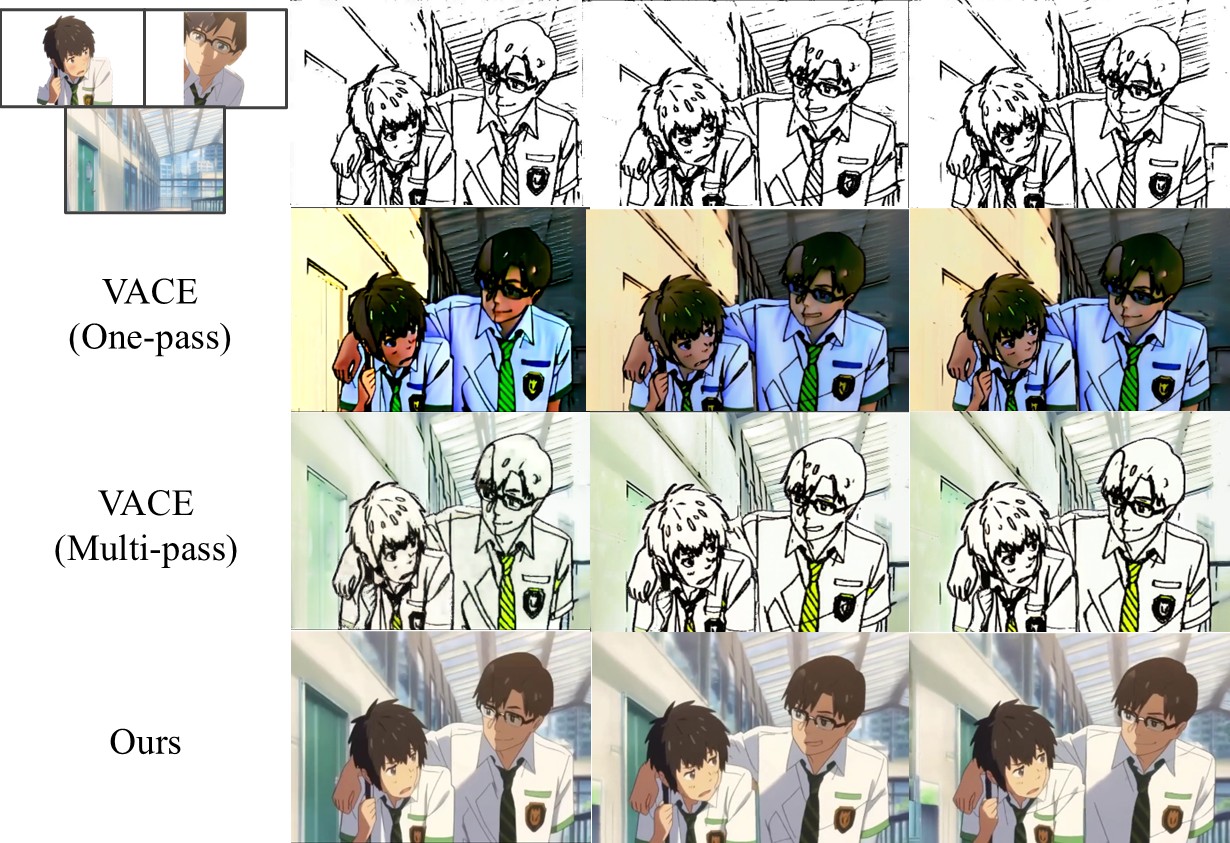}
\caption{\textbf{VACE multi-pass masked inference visualization.}
VACE one-pass multi-reference inference (without explicit region mapping) is compared with a per-reference multi-pass protocol using one binary masked edit per reference followed by sequential compositing.}
\label{fig:vace_multipass}
\end{figure}

As shown in Table~\ref{tab:vace_multipass}, the one-pass and multi-pass variants yield comparable frame-wise metrics in some cases, but the multi-pass protocol substantially degrades video-level quality, with a drastic increase in FVD. As qualitatively shown in Fig.~\ref{fig:vace_multipass}, sequential compositing introduces noticeable color transfer failure. For this reason, we report VACE one-pass in the main comparison to avoid penalizing VACE with a degraded workaround. In contrast, TimeColor enforces explicit multi-reference concurrently in each diffusion step with region assignment control.

\begin{table}[tbp]
\setlength{\tabcolsep}{1pt}
\caption{\textbf{VACE multi-pass masked inference (multi-reference).}
VACE one-pass multi-reference inference (no explicit region mapping) is compared with a per-reference multi-pass protocol that performs one binary masked edit per reference and sequentially composites the outputs.}
\label{tab:vace_multipass}
\centering
\begin{tabular}{@{}lrrrrr@{}}
\hline
Method & SSIM $\uparrow$ & PSNR $\uparrow$ & LPIPS $\downarrow$ & FVD $\downarrow$ & FID $\downarrow$ \\
\hline
\multicolumn{6}{c}{Multi-Reference} \\
\hline
VACE (One-pass)  & 0.3369 & 9.76 & 0.5342 & 454.40 & 64.01 \\
VACE (Multi-pass per Reference)    & 0.3485 & 8.63 & 0.5492 & 1054.42 & 79.51 \\
\textbf{TimeColor (Ours)} & \textbf{0.7589} & \textbf{18.89} & \textbf{0.2361} & \textbf{47.13} & \textbf{15.62}   \\
\hline
\end{tabular}
\end{table}

\subsection{Multi-Reference Evaluation via Tiled Reference Collage}
\label{app:TiledReferenceInput}
\begin{figure}[tbp]
\centering
\includegraphics[width=\linewidth]{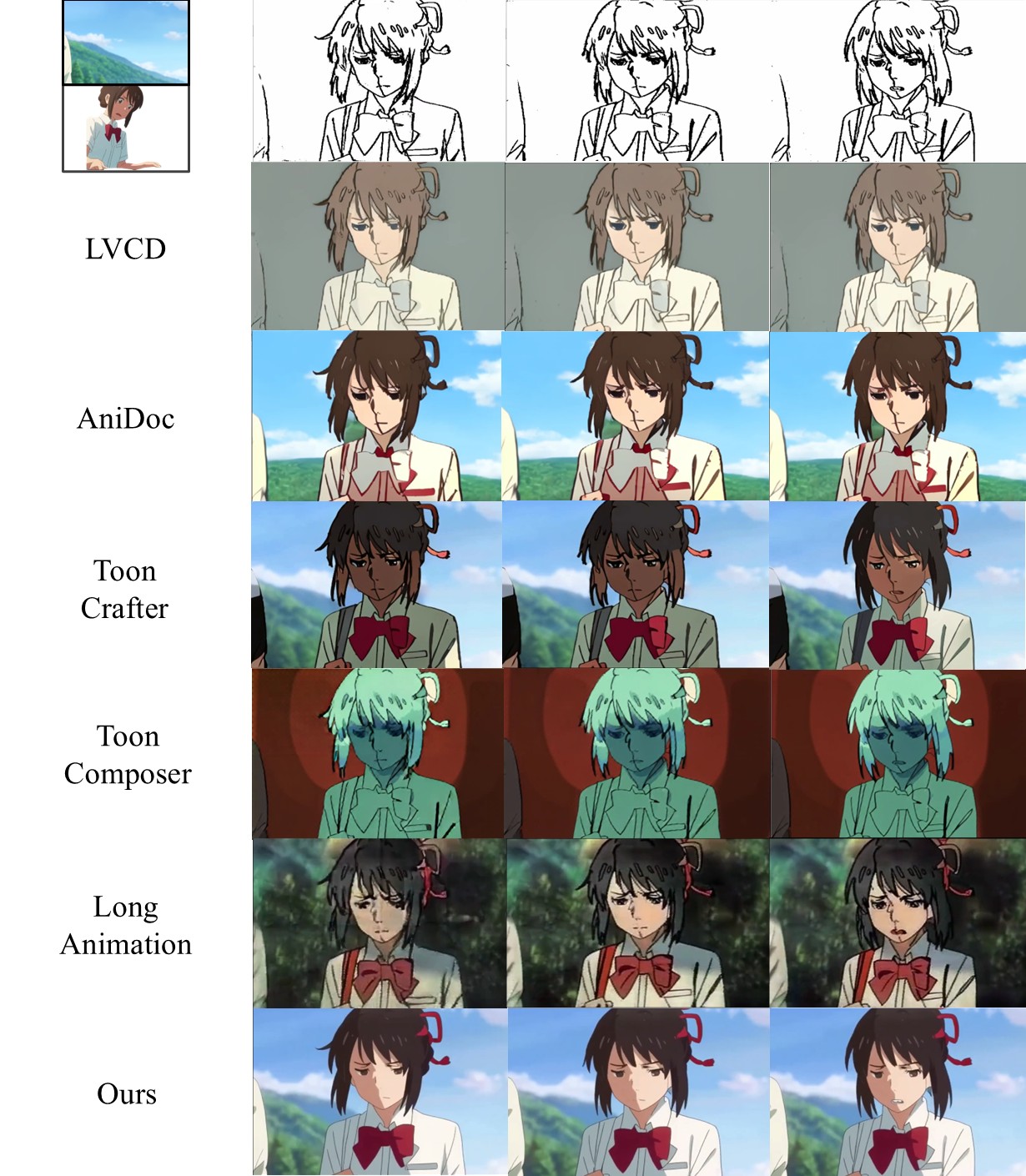}
\caption{\textbf{Alternative baseline multi-reference evaluation with tiled reference collages}. Multiple reference images are tiled into a single collage to enable baselines that only accept one colored reference input.}
\label{fig:qualitative_stage2_appendix}
\end{figure}

As an alternative multi-reference evaluation for baselines that only accept a single colored image, we follow AniDoc by tiling multiple reference images into a collage, which is then provided as the single reference input. As shown in Table~\ref{tab:tiled_multiref} and Fig.~\ref{fig:qualitative_stage2_appendix}, TimeColor achieves stronger color palette adherence to both subject and background references than this tiled-collage adaptation.

\begin{table}[tb]
\setlength{\tabcolsep}{3pt}
\caption{\textbf{Alternative multi-reference evaluation with tiled reference collages.} For baselines that accept only a single colored reference image, we tile multiple references into a collage and feed it as the input (following AniDoc). }
\label{tab:tiled_multiref}
\centering
\begin{tabular}{@{}lrrrrr@{}}
\hline
Method & SSIM $\uparrow$ & PSNR $\uparrow$ & LPIPS $\downarrow$ & FVD $\downarrow$ & FID $\downarrow$ \\
\hline
\multicolumn{6}{c}{Multi-Reference} \\
\hline
LVCD (Tiled Input)  & 0.4781 & 10.16 & 0.5454 & 618.22 & 72.81 \\
AniDoc  (Tiled Input)  & 0.5180 & 11.76 & 0.4257 & 423.21 & 59.74 \\
ToonCrafter (Tiled Input)   & 0.2341 & 9.47 & 0.5657 & 280.43 & 51.70 \\
ToonComposer (Tiled Input)   & 0.3096 & 9.61 & 0.6119 & 244.94 & 42.03 \\
LongAnimation (Tiled Input)    & 0.3780 & 10.88 & 0.5373 & 275.22 & 52.35 \\
\textbf{TimeColor (Ours)} & \textbf{0.7589} & \textbf{18.89} & \textbf{0.2361} & \textbf{47.13} & \textbf{15.62}   \\
\hline
\end{tabular}
\end{table}

\subsection{Alternative Arbitrary-Frame Reference Evaluation via Two-Step Colorization}
\label{app:ArbitraryCobraColorizer}
\begin{figure}[tbp]
\centering
\includegraphics[width=\linewidth]{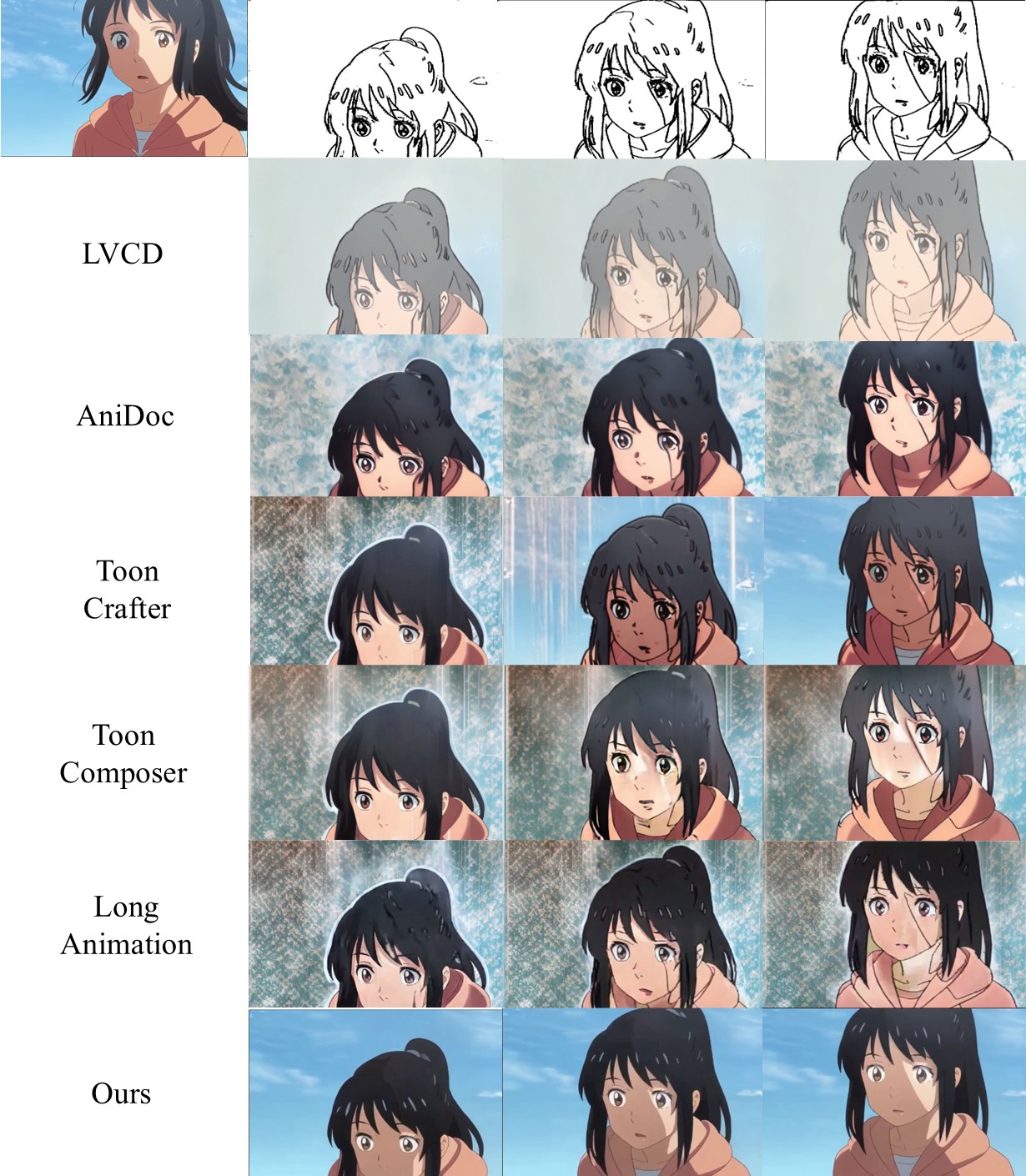}
\caption{\textbf{Alternative arbitrary-frame evaluation via a two-step pipeline}. We colorize the first frame with Cobra using an arbitrary-frame reference, then use the colorized first frame as the single-reference input to each baseline.}
\label{fig:qualitative_stage1_appendix}
\end{figure}

We also report an alternative protocol for arbitrary-frame references when benchmarking baselines. Specifically, a two-step pipeline is adopted: an image-to-image colorization model Cobra first colorizes the target clip’s starting-frame using the selected arbitrary-frame reference, and the resulting colored first frame is then used as the baseline’s single colored reference to colorize the remaining frames. As shown in Table~\ref{tab:ArbitraryCobraColorizer} and Fig.~\ref{fig:qualitative_stage1_appendix}, TimeColor follows the starting-frame reference guidance more faithfully, while avoiding the error accumulation introduced by the two-step inference pipeline.

\begin{table}[tbp]
\setlength{\tabcolsep}{3pt}
\caption{\textbf{Alternative baseline arbitrary-frame evaluations with additional image colorization model among baselines.}}
\label{tab:ArbitraryCobraColorizer}
\centering
\begin{tabular}{@{}lrrrrr@{}}
\hline
Method & SSIM $\uparrow$ & PSNR $\uparrow$ & LPIPS $\downarrow$ & FVD $\downarrow$ & FID $\downarrow$ \\
\hline
\multicolumn{6}{c}{Single Reference, Arbitrary-Frame} \\
\hline
LVCD (+ Cobra)  & 0.4874 & 9.73 & 0.5066 & 416.58 & 54.88 \\
AniDoc (+ Cobra)  & 0.6235 & 15.10 & 0.3609 & 135.73 & 36.56 \\
ToonCrafter (+ Cobra)   & 0.5585 & 14.43 & 0.3905 & 181.33 & 43.44 \\
ToonComposer (+ Cobra)  & 0.4714 & 12.86 & 0.4773 & 140.40 & 27.50 \\
LongAnimation (+ Cobra)    & 0.5152 & 13.75 & 0.4479 & 180.99 & 35.25 \\
\textbf{TimeColor (Ours)}                & \textbf{0.8071} & \textbf{21.98} & \textbf{0.1822} & \textbf{37.68} & \textbf{11.99}  \\
\hline
\end{tabular}
\end{table}

\subsection{Rationale for Sparse Sketch Selection in ToonComposer}
\label{app:ToonComposerSparseSketch}
\begin{figure}[tbp]
    \centering
    \includegraphics[width=\linewidth]{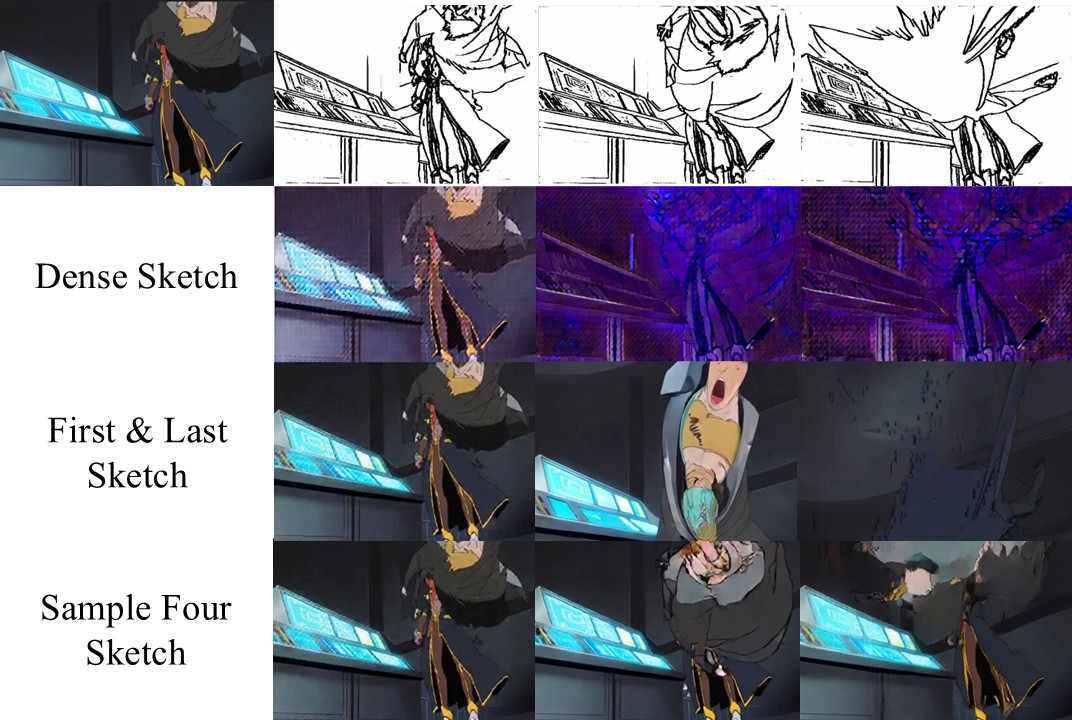}
    \caption{\textbf{ToonComposer samples under different sketch-selection strategies}. Dense sketches introduce noisy results, whereas four uniformly sampled sketches better preserve adherence to the input sketches.}
    \label{fig:tooncomposer_generation_comparison}
\end{figure}

As shown in Fig.~\ref{fig:tooncomposer_generation_comparison}, dense sketch conditioning of ToonComposer produces noisy results. We therefore adopt a sparse strategy in our main paper evaluation section to avoid penalizing ToonComposer with noisy output. Specifically, we evaluate two sparse sketch-sampling schemes: first/last only, and four uniformly spaced sketches (first/last + two in-between sketches) consistent with the authors’ Gradio demo configuration that uses at most four sketches. Table~\ref{tab:ToonComposerQuantAnalysis} shows that using four uniformly sampled sketches yields clear improvements in SSIM, PSNR, and LPIPS for starting-frame setting, with only slight changes in FID and FVD. Qualitatively (Fig.~\ref{fig:tooncomposer_generation_comparison}), four uniformly sampled sketches keep generations closer to the input sketches. Therefore, we apply this sampling scheme as our main evaluation protocol for ToonComposer.

\begin{table}[!tbp]
\caption{\textbf{Quantitative comparison of ToonComposer sketch-selection strategies.} "First\,+\,Last" indicates inference where only the first and last sketch are extracted, whereas "Four Uniform" samples four uniformly spaced sketches across the evaluation set.}
\label{tab:ToonComposerQuantAnalysis}
\centering
\begin{tabular}{@{}lrrrrr@{}}
\hline
Sketch Selection & SSIM $\uparrow$ & PSNR $\uparrow$ & LPIPS $\downarrow$ & FVD $\downarrow$ & FID $\downarrow$ \\
\hline
\multicolumn{6}{c}{Single Reference, Starting-Frame} \\
\hline
First \& Last & 0.6519 & 18.35 & 0.3197 & 65.28 & 11.96 \\
Four Uniform  & 0.7046 & 20.09 & 0.2371 & 67.54 & 12.78 \\
\hline
\end{tabular}
\end{table}

